\documentclass[10pt,twocolumn,letterpaper]{article}

\usepackage{iccv}
\usepackage{times}
\usepackage{epsfig}
\usepackage{graphicx}
\usepackage{amsmath}
\usepackage{amssymb}

\usepackage[caption=false]{subfig}
\usepackage{mathrsfs}
\usepackage{mathtools}
\usepackage{booktabs}
\usepackage{array}
\usepackage{cases}
\usepackage{mynotation}

\usepackage[breaklinks=true,bookmarks=false,colorlinks=true]{hyperref}

\iccvfinalcopy % *** Uncomment this line for the final submission

\def\iccvPaperID{1101} % *** Enter the ICCV Paper ID here

\ificcvfinal\fi
\begin{document}

%%%%%%%%% TITLE
\title{Learning Spatially Regularized Correlation Filters for Visual Tracking}

\author{ Martin Danelljan, Gustav H\"ager, Fahad Shahbaz Khan, Michael Felsberg \\
	\small Computer Vision Laboratory, Link\"oping University, Sweden\\
	\small\{\texttt{martin.danelljan},\; \texttt{gustav.hager},\; \texttt{fahad.khan},\; \texttt{michael.felsberg}\}\texttt{@liu.se}
	{}
}

\maketitle
\thispagestyle{empty}

%%%%%%%%% ABSTRACT
\begin{abstract}
	Robust and accurate visual tracking is one of the most challenging computer vision problems. Due to the inherent lack of training data, a robust approach for constructing a target appearance model is crucial. Recently, discriminatively learned correlation filters (DCF) have been successfully applied to address this problem for tracking. These methods utilize a periodic assumption of the training samples to efficiently learn a classifier on all patches in the target neighborhood. However, the periodic assumption also introduces unwanted boundary effects, which severely degrade the quality of the tracking model.

We propose Spatially Regularized Discriminative Correlation Filters (SRDCF) for tracking. A spatial regularization component is introduced in the learning to penalize correlation filter coefficients depending on their spatial location. Our SRDCF formulation allows the correlation filters to be learned on a significantly larger set of negative training samples, without corrupting the positive samples. We further propose an optimization strategy, based on the iterative Gauss-Seidel method, for efficient online learning of our SRDCF. Experiments are performed on four benchmark datasets: OTB-2013, ALOV++, OTB-2015, and VOT2014. Our approach achieves state-of-the-art results on all four datasets. On OTB-2013 and OTB-2015, we obtain an absolute gain of $8.0 \%$ and $8.2 \%$ respectively, in mean overlap precision, compared to the best existing trackers.
\end{abstract}

%%%%%%%%% BODY TEXT
\section{Introduction}

Visual tracking is a classical computer vision problem with many applications. In generic tracking the task is to estimate the trajectory of a target in an image sequence, given only its initial location. This problem is especially challenging. The tracker must generalize the target appearance from a very limited set of training samples to achieve robustness against, \eg occlusions, fast motion and deformations. Here, we investigate the key problem of learning a robust appearance model under these conditions.

\begin{figure}[!t]
	\newcommand{\wid}{0.23\textwidth}
	\newcommand{\trimh}{30}
	\centering\vspace{-1mm}
	\subfloat[Original image.]{\includegraphics*[trim=0 0 0 0, width=\wid]{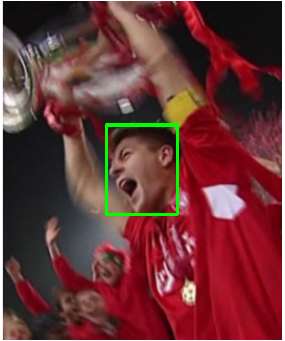}}\hspace{0.1mm}
	\subfloat[Periodicity in correlation filters.]{\includegraphics*[trim= 0 0 0 0, width=\wid]{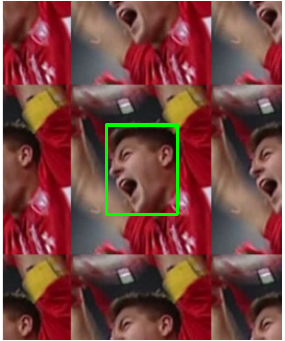}}\vspace{1mm}
	\caption{Example image (a) and the underlying periodic assumption (b) employed in the standard DCF methods. The periodic assumption (b) leads to a limited set of negative training samples, that fails to capture the true image content (a). As a consequence, an inaccurate tracking model is learned.}
	\vspace{-4mm}
	\label{fig:periodic}
\end{figure}

Recently, Discriminative Correlation Filter (DCF) based approaches \cite{MOSSE2010,DanelljanBMVC14,DanelljanCVPR14,Henriques12d,HenriquesPAMI15,Li2014} have successfully been applied to the tracking problem \cite{VOT2014}. These methods learn a correlation filter from a set of training samples. The correlation filter is trained to perform a circular sliding window operation on the training samples. This corresponds to assuming a periodic extension of these samples (see figure~\ref{fig:periodic}). The periodic assumption enables efficient training and detection by utilizing the Fast Fourier Transform (FFT). 

As discussed above, the computational efficiency of the standard DCF originates from the periodic assumption at both training and detection. However, this underlying assumption produces unwanted boundary effects. This leads to an inaccurate representation of the image content, since the training patches contain periodic repetitions. The induced boundary effects mainly limit the standard DCF formulation in two important aspects. Firstly, inaccurate negative training patches reduce the discriminative power of the learned model. Secondly, the detection scores are only accurate near the center of the region, while the remaining scores are heavily influenced by the periodic repetitions of the detection sample. This leads to a very restricted target search region at the detection step. 

The aforementioned limitations of the standard DCF formulation hamper the tracking performance in several ways. (a) The DCF based trackers struggle in cases with fast target motion due to the restricted search region. (b) The lack of negative training patches leads to over-fitting of the learned model, significantly affecting the performance in cases with \eg target deformations. (c) The mentioned limitations in training and detection also reduce the potential of the tracker to re-detect the target after an occlusion. (d)  A naive expansion of the image area used for training the correlation filter corresponds to using a larger periodicity (see figure~\ref{fig:periodic}). Such an expansion results in an inclusion of substantial amount of background information within the positive training samples. These corrupted training samples severely degrade the discriminative power of the model, leading to inferior tracking results.  In this work, we tackle these inherent problems by re-visiting the standard DCF formulation.

\subsection{Contributions}
In this paper, we propose Spatially Regularized Discriminative Correlation Filters (SRDCF) for tracking. We introduce a spatial regularization component within the DCF formulation, to address the problems induced by the periodic assumption. The proposed regularization weights penalize the correlation filter coefficients during learning. The spatial weights are based on the a priori information about the spatial extent of the filter. Due to the spatial regularization, the correlation filter can be learned on larger image regions. This enables a larger set of negative patches to be included in the training, leading to a more discriminative model.

Due to the online nature of the tracking problem, a computationally efficient learning scheme is crucial. Therefore, we introduce a suitable optimization strategy for the proposed SRDCF. The online capability is achieved by exploiting the sparsity of the spatial regularization function in the Fourier domain. We propose to apply the iterative Gauss-Seidel method to solve the resulting normal equations. 
Additionally, we introduce a strategy to maximize the detection scores with sub-grid precision.

We perform comprehensive experiments on four benchmark datasets: OTB-2013 \cite{Wu13g} with 50 videos, ALOV++ \cite{ALOV} with 314 videos, VOT2014 \cite{VOT2014} with 25 videos and OTB-2015 \cite{OTB2015} with 100 videos. Compared to the best existing trackers, our approach obtains an absolute gain of $8.0 \%$ and $8.2 \%$ on OTB-2013 and OTB-2015 respectively, in mean overlap precision. Our method also achieves the best overall results on ALOV++ and VOT2014. Additionally, our tracker won the OpenCV State of the Art Vision Challenge in tracking \cite{OpenCVchallenge} (there termed DCFSIR).

\section{Discriminative Correlation Filters}

Discriminative correlation filters (DCF) is a supervised technique for learning a linear classifier or a linear regressor. The main difference from other techniques, such as support vector machines \cite{VapnikSVM}, is that the DCF formulation exploits the properties of circular correlation for efficient training and detection. In recent years, the DCF based approaches have been successfully applied for tracking. Bolme \etal \cite{MOSSE2010} first introduced the MOSSE tracker, using only grayscale samples to train the filter.
Recent work \cite{DanelljanSCIA2015,DanelljanBMVC14,DanelljanCVPR14,HenriquesPAMI15,Li2014} have shown a notable improvement by learning multi-channel filters on multi-dimensional features, such as HOG \cite{Dalal05} or Color-Names \cite{Weijer09a}. However, to become computationally viable, these approaches rely on harsh approximations of the standard DCF formulation, leading to sub-optimal learning. Other work have investigated offline learning of multi-channel DCF:s for object detection \cite{galoogahiICCV13,henriquesICCV13} and recognition \cite{BoddetiCVPR13}. But these methods are too computationally costly for online tracking applications.

The circular correlation within the DCF formulation has two major advantages. Firstly, the DCF is able to make extensive use of limited training data by implicitly including all shifted versions of the given samples. Secondly, the computational effort for training and detection is significantly reduced by performing the necessary computations in the Fourier domain and using the Fast Fourier Transform (FFT). These two advantages make DCF:s especially suitable for tracking, where training data is scarce and computational efficiency is crucial for real-time applications. 

By employing a circular correlation, the standard DCF formulation relies on a periodic assumption of the training and detection samples. However, this assumption produces unwanted boundary effects, leading to an inaccurate description of the image. These inaccurate training patches severely hamper the learning of a discriminative tracking model. Surprisingly, this problem has been largely ignored by the tracking community. Galoogahi \etal \cite{GaloogahiCVPR2015} investigate the boundary effect problem for single-channel DCF:s. Their approach solve a constrained optimization problem, using the Alternating Direction Method of Multipliers (ADMM), to ensure a correct filter size. This however requires a transition between the spatial and Fourier domain in each ADMM iteration, leading to an increased computational complexity. Different to \cite{GaloogahiCVPR2015}, we propose a spatial regularization component in the objective. By exploiting the sparsity of our regularizer, we efficiently optimize the filter directly in the Fourier domain. Contrary to \cite{GaloogahiCVPR2015}, we target the problem of multi-dimensional features, such as HOG, crucial for the overall tracking performance \cite{DanelljanCVPR14,HenriquesPAMI15}.

\subsection{Standard DCF Training and Detection}
In the DCF formulation, the aim is to learn a multi-channel convolution\footnote{We use convolution for mathematical convenience, though correlation can equivalently be used.} filter $f$ from a set of training examples $\{(x_k, y_k)\}_{k=1}^t$. Each training sample $x_k$ consists of a $d$-dimensional feature map extracted from an image region. All samples are assumed to have the same spatial size $M \times N$. At each spatial location $(m,n) \in \Omega \defeq \{0,\ldots,M-1\} \times \{0,\ldots,N-1\}$ we thus have a $d$-dimensional feature vector $x_k(m,n) \in \reals^d$. We denote feature layer $l \in \{1,\ldots,d\}$ of $x_k$ by $x_k^l$. The desired output $y_k$ is a scalar valued function over the domain $\Omega$, which includes a label for each location in the sample $x_k$.

The desired filter $f$ consists of one $M \times N$ convolution filter $f^l$ per feature layer. The convolution response of the filter $f$ on a $M \times N$ sample $x$ is given by
\begin{equation}
	\label{eq:mdcf_conv}
	S_f(x) = \sum_{l=1}^d x^l \conv f^l .
\end{equation}
Here, $\conv$ denotes circular convolution. The filter is obtained by minimizing the $L^2$-error between the responses $S_f(x_k)$ on the training samples $x_k$, and the labels $y_k$,
\begin{equation}
	\label{eq:mdcf_cost}
	\varepsilon_t(f) = \sum_{k=1}^t \alpha_k \big\| S_f(x_k) - y_k \big\|^2 + \lambda \sum_{l=1}^d \big\| f^l \big\|^2 .
\end{equation}
Here, the weights $\alpha_k \geq 0$ determine the impact of each training sample and $\lambda \geq 0$ is the weight of the regularization term. Eq.~\ref{eq:mdcf_cost} is a linear least squares problem. Using Parseval's formula, it can be transformed to the Fourier domain, where the resulting normal equations have a block diagonal structure. The Discrete Fourier Transformed (DFT) filters $\hat{f}^l = \ft\{f^l\}$ can then be obtained by solving $MN$ number of $d \times d$ linear equation systems \cite{galoogahiICCV13}. 

For efficiency reasons, the learned DCF is typically applied in a sliding-window-like manner by evaluating the classification scores on all cyclic shifts of a test sample. Let $z$ denote the $M \times N$ feature map extracted from an image region. The classification scores $S_f(z)$ at all locations in this image region can be computed using the convolution property of the DFT,
\begin{equation}
	\label{eq:mdcf_detection}
	S_f(z) = \ftinv \Bigg\{ \sum_{l=1}^d \hat{z}^l \pmult \hat{f}^l \Bigg\} .
\end{equation}
Here, $\pmult$ denotes point-wise multiplication, the hat denotes the DFT of a function and $\ftinv$ denotes the inverse DFT. The FFT hence allows the detection scores to be computed in $\ordo(dMN \log MN)$ complexity instead of $\ordo(dM^2N^2)$.

Note that the operation $S_f(x)$ in \eqref{eq:mdcf_conv} corresponds to applying the linear classifier $f$, in a sliding window fashion, to the periodic extension of the sample $x$ (see figure~\ref{fig:periodic}). This introduces unwanted periodic boundary effects in the training \eqref{eq:mdcf_cost} and detection \eqref{eq:mdcf_detection} steps.

\section{Spatially Regularized Correlation Filters}
\label{sec:method}
We propose to use a spatial regularization component in the standard DCF formulation. The resulting optimization problem is solved in the Fourier domain, by exploiting the sparse nature of the proposed regularization. 

\begin{figure}[!t]
	\newcommand{\wid}{0.46\textwidth}
	\centering
	\includegraphics[width=\wid,height=5cm]{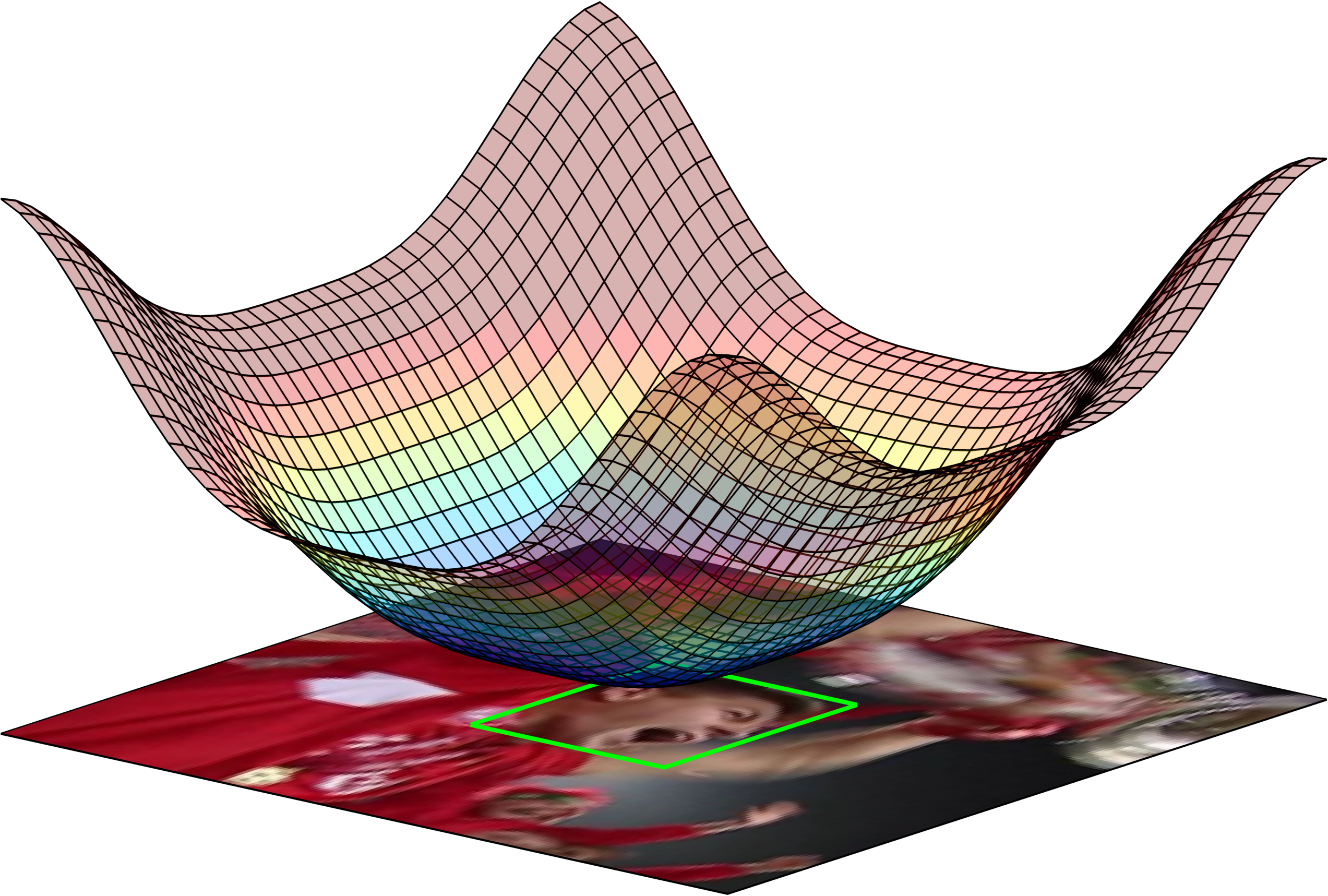}\vspace{-0.5mm}
	\caption{Visualization of the spatial regularization weights $w$ employed in the learning of our SRDCF, and the corresponding image region used for training. Filter coefficients residing in the background region are penalized by assigning higher weights in $w$. This significantly mitigates the emphasis on background information in the learned classifier.}
	\vspace{-1mm}
	\label{fig:weights}
\end{figure}

\subsection{Spatial Regularization}
To alleviate the problems induced by the circular convolution in \eqref{eq:mdcf_conv}, we replace the regularization term in \eqref{eq:mdcf_cost} with a more general Tikhonov regularization. We introduce a spatial weight function $w: \Omega \rightarrow \reals$ used to penalize the magnitude of the filter coefficients in the learning.
The regularization weights $w$ determine the importance of the filter coefficients $f^l$, depending on their spatial locations. Coefficients in $f^l$ residing outside the target region are suppressed by assigning higher weights in $w$ and vice versa.
The resulting optimization problem is expressed as,
\begin{equation}
	\label{eq:ours_cost_spatial}
	\varepsilon(f) = \sum_{k=1}^t \alpha_k \big\| S_f(x_k) - y_k \big\|^2 + \sum_{l=1}^d \big\| w \pmult f^l \big\|^2 .
\end{equation}
The regularization weights $w$ in \eqref{eq:ours_cost_spatial} are visualized in figure~\ref{fig:weights}. Visual features close to the target edge are often less reliable than those close to the target center, due to \eg target rotations and occlusions. We therefore let the regularization weights change smoothly from the target region to the background. This also increases the sparsity of $w$ in the Fourier domain.
Note that \eqref{eq:ours_cost_spatial} simplifies to the standard DCF \eqref{eq:mdcf_cost} for uniform weights $w(m,n) = \sqrt{\lambda}$.

\begin{figure*}[!t]
	\newcommand{\wid}{0.49\textwidth}
	\centering
	\subfloat[Standard DCF.]{\includegraphics[width=\wid]{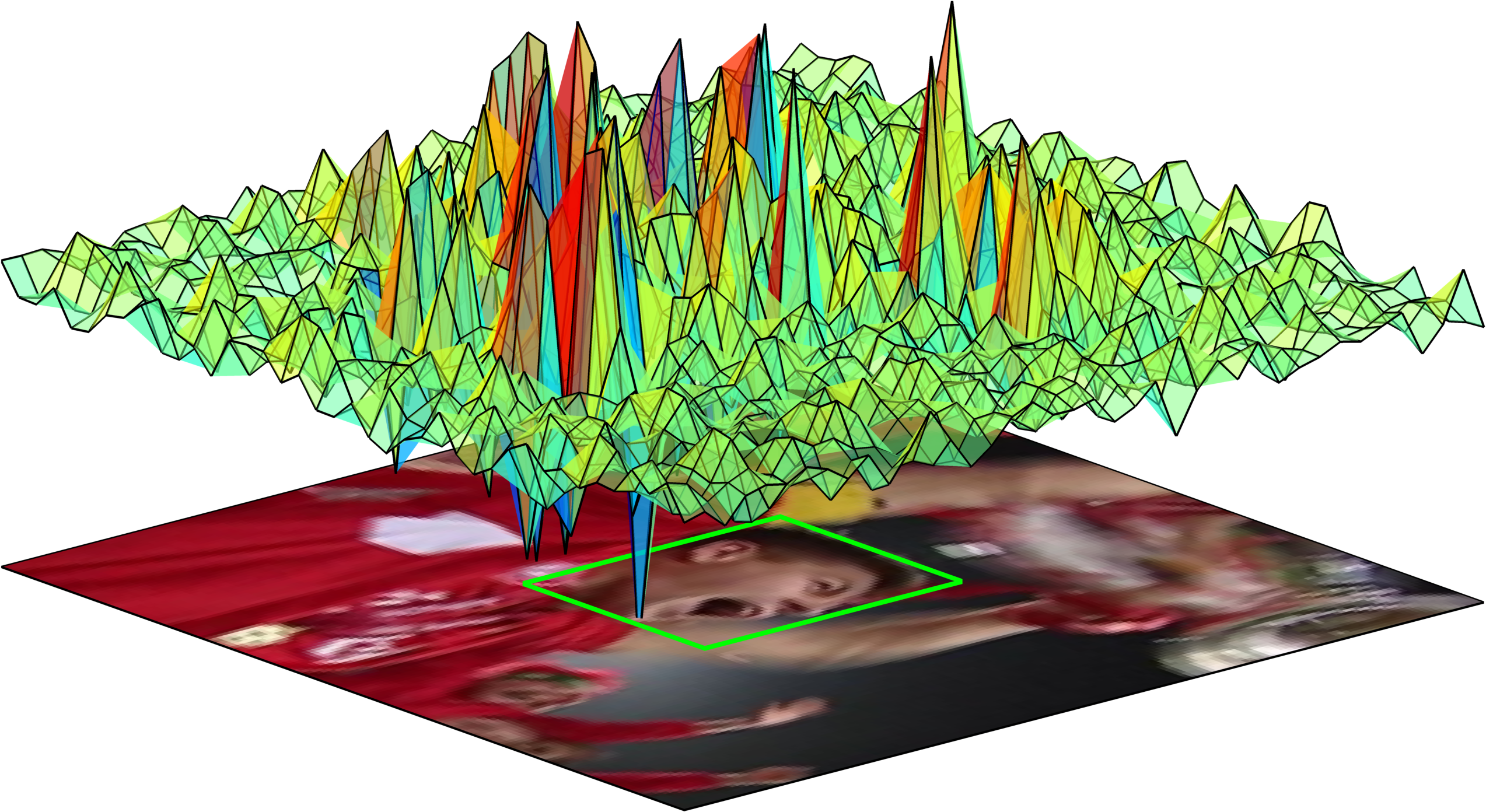}}\hspace{1mm}
	\subfloat[Our SRDCF.]{\includegraphics[width=\wid]{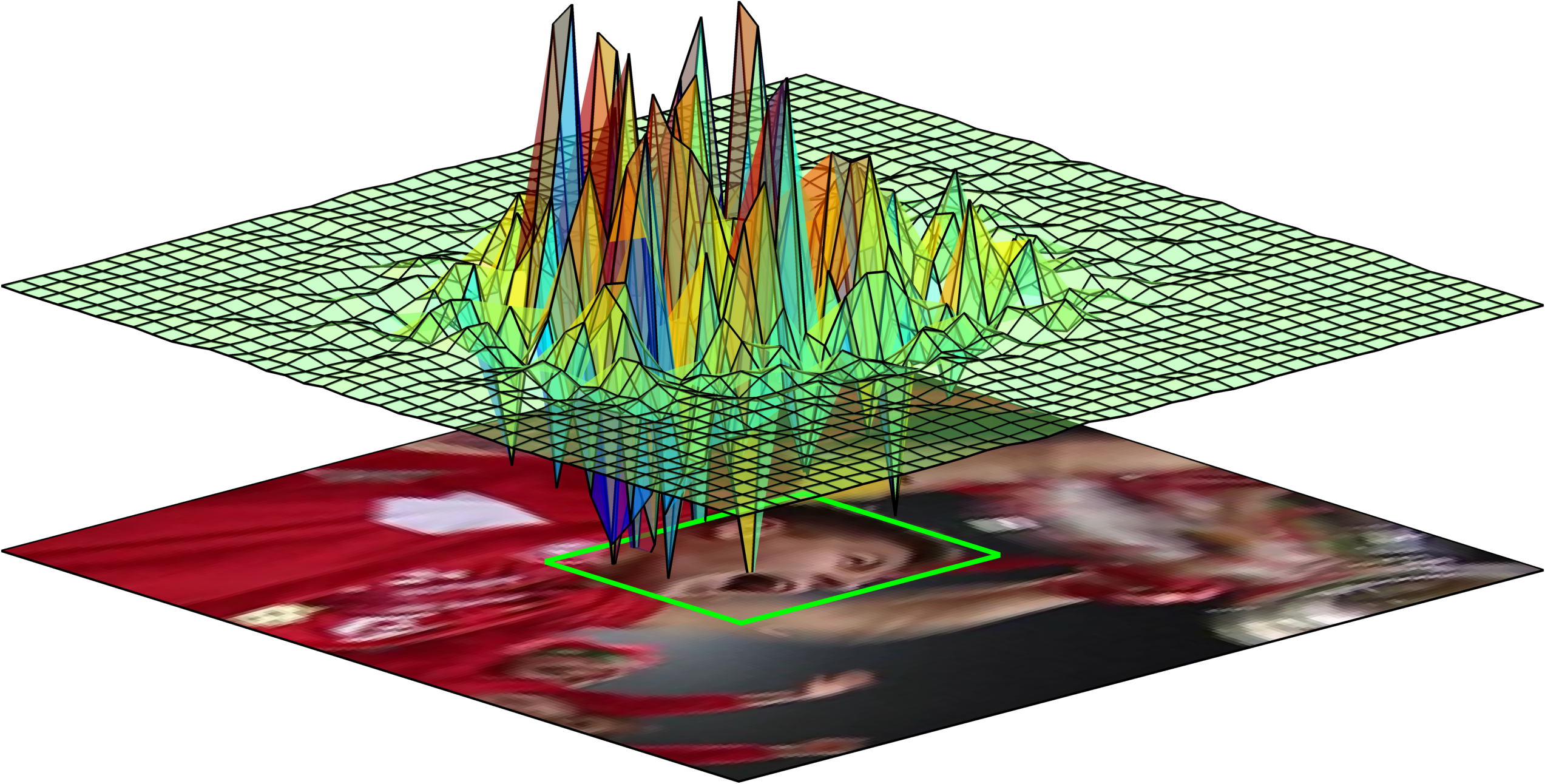}}\vspace{0.0mm}
	\caption{Visualization of the filter coefficients learned using the standard DCF (a) and our approach (b). The surface plots show the filter values $f^l$ and the corresponding image region used for training. In the standard DCF, high values are assigned to the background region. The larger influence of background information at the detection stage deteriorates tracking performance. In our approach, the regularization weights penalizes filter values corresponding to features in the background. This increases the discriminative power of the learned model, by emphasizing the appearance information within the target region (green box).}
	\vspace{-4mm}
	\label{fig:filters}
\end{figure*}

By applying Parseval's theorem to \eqref{eq:ours_cost_spatial}, the filter $f$ can equivalently be obtained by minimizing the resulting loss function \eqref{eq:ours_cost_dft} over the DFT coefficients $\hat{f}$,
\begin{equation}
	\label{eq:ours_cost_dft}
	\check{\varepsilon}(\hat{f}) = \sum_{k=1}^t \alpha_k \Bigg\| \sum_{l=1}^d \hat{x}_k^l \pmult \hat{f}^l - \hat{y}_k \Bigg\|^2 + \sum_{l=1}^d \Bigg\| \frac{\hat{w}}{MN} \conv \hat{f}^l \Bigg\|^2 .
\end{equation}
The second term in \eqref{eq:ours_cost_dft} follows from the convolution property of the inverse DFT.
A vectorization of \eqref{eq:ours_cost_dft} gives,
\begin{equation}
	\label{eq:ours_cost_vec1}
	\check{\varepsilon}(\hat{f}) = \sum_{k=1}^t \alpha_k \Bigg\| \sum_{l=1}^d \diagm(\vecn{\hat{x}}_k^l) \vecn{\hat{f}}^l - \vecn{\hat{y}}_k \Bigg\|^2 \hspace{-1.0mm} + \sum_{l=1}^d \Bigg\| \frac{\circm(\vecn{\hat{w}})}{MN} \vecn{\hat{f}}^l \Bigg\|^2 \hspace{-1.0mm}.
\end{equation}
Here, bold letters denote a vectorization of the corresponding scalar valued functions and $\diagm(\vecn{v})$ denotes the diagonal matrix with the elements of the vector $\vecn{v}$ in its diagonal. The $MN \times MN$ matrix $\circm(\vecn{\hat{w}})$ represents circular 2D-convolution with the function $\hat{w}$, \ie $\circm(\vecn{\hat{w}}) \vecn{\hat{f}}^l = \vecnf(\hat{w} \conv \hat{f}^l)$. Each row in $\circm(\vecn{\hat{w}})$ thus contains a cyclic permutation of $\vecn{\hat{w}}$.

The DFT of a real-valued function is known to be Hermitian symmetric. Therefore, minimizing \eqref{eq:ours_cost_spatial} over the set of real-valued filters $f^l$, corresponds to minimizing \eqref{eq:ours_cost_dft} over the set of Hermitian symmetric DFT coefficients $\hat{f}^l$. We reformulate \eqref{eq:ours_cost_vec1} to an equivalent real-valued optimization problem, to ensure faster convergence by preserving the Hermitian symmetry. Let $\rho : \Omega \rightarrow \Omega$ be the point-reflection $\rho(m,n) = (-m \mod M,-n \mod N)$. The domain $\Omega$ can be partitioned into $\Omega_0$, $\Omega_+$ and $\Omega_-$, where $\Omega_0 = \rho(\Omega_0)$ and $\Omega_- = \rho(\Omega_+)$. Thus, $\Omega_0$ denote the part of the spectrum with no corresponding reflected frequency, and  $\Omega_-$ contains the reflected frequencies in $\Omega_+$. We define,
\begin{equation}
	\label{eq:real_trans}
	\tilde{f}^l(m,n) =
	\begin{cases}
	    \hat{f}^l(m,n) , & (m,n) \in \Omega_0 \\
		\frac{\hat{f}^l(m,n) + \hat{f}^l(\rho(m,n))}{\sqrt{2}} , & (m,n) \in \Omega_+ \\
		\frac{\hat{f}^l(m,n) - \hat{f}^l(\rho(m,n))}{i\sqrt{2}} , & (m,n) \in \Omega_-
	\end{cases}
\end{equation}
such that  $\tilde{f}^l$ is real-valued by the Hermitian symmetry of $\hat{f}^l$. Here, $i$ denotes the imaginary unit. Eq.~\ref{eq:real_trans} can be expressed by a unitary $MN \times MN$ matrix $B$ such that $\vectil{f}^l = B\vecft{f}^l$. By \eqref{eq:real_trans}, $B$ contains at most two non-zero entries in each row.

The reformulated variables from \eqref{eq:ours_cost_vec1} are defined as $\vectil{y}_k = B \vecft{y}_k$, $\redftmat{D}_k^l = B \diagm(\vecn{\hat{x}}_k^l) B\ctp$ and $\redftmat{C} = \frac{1}{MN} B \circm(\vecn{\hat{w}}) B\ctp$,
where $\ctp$ denotes the conjugate transpose of a matrix. Since $B$ is unitary, \eqref{eq:ours_cost_vec1} can equivalently be expressed as,
\begin{equation}
\label{eq:ours_cost_vec2}
\tilde{\varepsilon}(\vectil{f}^1\ldots\vectil{f}^d) = \sum_{k=1}^t \alpha_k \Bigg\| \sum_{l=1}^d \redftmat{D}_k^l \vectil{f}^l - \vectil{y}_k \Bigg\|^2 \hspace{-2mm} + \sum_{l=1}^d \left\| \redftmat{C} \vectil{f}^l \right\|^2 \hspace{-2mm}.
\end{equation}
All variables in \eqref{eq:ours_cost_vec2} are real-valued. The loss function \eqref{eq:ours_cost_vec2} is then simplified by defining the fully vectorized real-valued filter as the concatenation $\vectil{f} = \big((\vectil{f}^1)\tp \cdots (\vectil{f}^d)\tp \big)\tp$,
\begin{equation}
	\label{eq:ours_cost_vec3}
	\tilde{\varepsilon}(\vectil{f}) = \sum_{k=1}^t \alpha_k \Big\| \redftmat{D}_k \vectil{f} - \vectil{y}_k \Big\|^2 + \Big\| \redftmat{W} \vectil{f} \Big\|^2 .
\end{equation}
Here we have defined the concatenation $\redftmat{D}_k = (\redftmat{D}_k^1 \cdots \redftmat{D}_k^d)$ and $\redftmat{W}$ to be the $dMN \times dMN$ block diagonal matrix with each diagonal block being equal to $\redftmat{C}$. Finally, \eqref{eq:ours_cost_vec3} is minimized by solving the normal equations $\redftmat{A}_t \vectil{f} = \vectil{b}_t$, where%
\begin{subequations}
	\label{eq:normal_equations_real}
	\begin{align}
	\label{eq:normal_A_real}
	\redftmat{A}_t &= \sum_{k=1}^t \alpha_k \redftmat{D}_k\tp \redftmat{D}_k + \redftmat{W}\tp \redftmat{W} \\
	\tilde{\mathbf{b}}_t &= \sum_{k=1}^t \alpha_k \redftmat{D}_k\tp \tilde{\vecn{y}}_k .
	\end{align}
\end{subequations}
Here, \eqref{eq:normal_equations_real} defines a real $dMN \times dMN$ linear system of equations. The fraction of non-zero elements in $\redftmat{A}_t$ is smaller than $\frac{2d + K^2}{dMN}$, where $K$ is the number of non-zero Fourier coefficients in $\hat{w}$. Thus, $\redftmat{A}_t$ is sparse if $w$ has a sparse spectrum. The DFT coefficients for the filters are obtained by solving the system \eqref{eq:normal_equations_real} and applying $\vecft{f}^l = B\ctp \vectil{f}^l$.

Figure~\ref{fig:filters} visualizes the filter learned by optimizing the standard DCF loss \eqref{eq:mdcf_cost} and the proposed formulation \eqref{eq:ours_cost_spatial}, using the spatial regularization weights $w$ in figure~\ref{fig:weights}. In the standard DCF, large values are spatially distributed over the whole filter. By penalizing filter coefficients corresponding to background, our approach learns a classifier that emphasizes visual information within the target region.

A direct application of a sparse solver to the normal equations $\redftmat{A}_t \vectil{f} = \vectil{b}_t$ is computationally very demanding (even when the standard regularization $\redftmat{W}\tp \redftmat{W} = \lambda I$ is used and the number of features is small ($d > 2$)). Next, we propose an efficient optimization scheme to solve the normal equations for online learning scenarios, such as tracking.

\subsection{Optimization}
For the standard DCF formulation \eqref{eq:mdcf_cost} the normal equations have a block diagonal structure \cite{galoogahiICCV13}. However, this block structure is not attainable in our case due to the structure of the regularization matrix $\redftmat{W}\tp \redftmat{W}$ in \eqref{eq:normal_A_real}. We propose an iterative approach, based on the Gauss-Seidel, for efficient online computation of the filter coefficients.

The Gauss-Seidel method decomposes the matrix $\redftmat{A}_t$ into a lower triangular part $\redftmat{L}_t$ and a strictly upper triangular part $\redftmat{U}_t$ such that $\redftmat{A}_t = \redftmat{L}_t + \redftmat{U}_t$. The algorithm then proceeds by solving the following triangular system for $\vectil{f}^{(j)}$ in each iteration $j = 1, 2, \ldots$,
\begin{equation}
	\label{eq:gauss_seidel}
	\redftmat{L}_t \vectil{f}^{(j)} = \tilde{\mathbf{b}}_t - \redftmat{U}_t \vectil{f}^{(j-1)} .
\end{equation}
This lower triangular equation system is solved efficiently using forward substitution and by exploiting the sparsity of $\redftmat{L}_t$ and $\redftmat{U}_t$. The Gauss-Seidel recursion \eqref{eq:gauss_seidel} converges to the solution of $\redftmat{A}_t \vectil{f} = \tilde{\mathbf{b}}$ whenever the matrix $\redftmat{A}_t$ is symmetric and positive definite. The construction of the weights $w$ (see section~\ref{sec:details}) ensures that both conditions are satisfied.

\section{Our Tracking Framework}
Here, we describe our tracking framework, based on the Spatially Regularized Discriminative Correlation Filters (SRDCF) proposed in section~\ref{sec:method}. 

\subsection{Training}
At the training stage, the model is updated by first extracting a new training sample $x_t$ centered at the target location. Here, $t$ denotes the current frame number. We then update $\redftmat{A}_t$ and $\tilde{\mathbf{b}}_t$ in \eqref{eq:normal_equations_real} with a learning rate $\gamma \geq 0$,%
\begin{subequations}
	\label{eq:normal_equations_real_update}
	\begin{align}
	\label{eq:normal_A_real_update}
	\redftmat{A}_t &= (1 - \gamma) \redftmat{A}_{t-1} + \gamma \left( \redftmat{D}_t\tp \redftmat{D}_t + \redftmat{W}\tp \redftmat{W} \right) \\
	\tilde{\mathbf{b}}_t &= (1 - \gamma) \tilde{\mathbf{b}}_{t-1} + \gamma \redftmat{D}_t\tp \tilde{\vecn{y}}_t .
	\end{align}
\end{subequations}
This corresponds to using exponentially decaying weights $\alpha_k$ in the loss function \eqref{eq:ours_cost_spatial}. In the first frame, we set $\redftmat{A}_1 = \redftmat{D}_1\tp \redftmat{D}_1 + \redftmat{W}\tp \redftmat{W}$ and $\tilde{\mathbf{b}}_1 = \redftmat{D}_1\tp \tilde{\vecn{y}}_1$. Note that the regularization matrix $\redftmat{W}\tp \redftmat{W}$ can be precomputed once for the entire sequence. The update strategy \eqref{eq:normal_equations_real_update} ensures memory efficiency, since it does not require storage of all samples $x_k$.  
After the model update \eqref{eq:normal_equations_real_update}, we perform a fixed number $N_\text{GS}$ of Gauss-Seidel iterations \eqref{eq:gauss_seidel} per frame to compute the new filter coefficients.

For the initial iteration $\vectil{f}_t^{(0)}$ in frame $t$, we use the filter computed in the previous frame, \ie $\vectil{f}_t^{(0)} = \vectil{f}_{t-1}^{(N_\text{GS})}$. In the first frame, the initial estimate $\vectil{f}_1^{(0)}$ is obtained by solving the $MN \times MN$ linear system,
\begin{equation}
	\label{eq:initial_solution}
	\left(\sum_{p=1}^d (\redftmat{D}_1^p)\tp \redftmat{D}_1^p + d \redftmat{C}\tp \redftmat{C} \right) \vectil{f}_1^{l,(0)} = (\redftmat{D}_1^l)\tp \tilde{\vecn{y}}_1
\end{equation}
for $l = 1,\ldots,d$. 
This provides a starting point for the Gauss-Seidel optimization in the first frame. The systems in \eqref{eq:initial_solution} share the same sparse coefficients and can be solved efficiently with a direct sparse solver.

\subsection{Detection}
At the detection stage, the location of the target in a new frame $t$ is estimated by applying the filter $\hat{f}_{t-1}$ that has been updated in the previous frame. Similar to \cite{Li2014}, we apply the filter at multiple resolutions to estimate changes in the target size. The samples $\{z_r\}_{r \in \left\{\left\lfloor\frac{1-S}{2}\right\rfloor, \ldots, \left\lfloor\frac{S-1}{2}\right\rfloor\right\}}$ are extracted centered at the previous target location and at the scales $a^r$ relative to the current target scale. Here, $S$ denotes the number of scales and $a$ is the scale increment factor. The sample $z_r$ is constructed by resizing the image according to $a^r$ before the feature computation.

\noindent \textbf{Fast Sub-grid Detection}:
Generally, the training and detection samples $x_k$ and $z_k$ are constructed using a grid strategy with a stride greater than one pixel. This leads to only computing the detection scores  \eqref{eq:mdcf_detection} on a coarser grid. We employ an interpolation approach that allows computation of pixel-dense detection scores. The detection scores \eqref{eq:mdcf_detection} are efficiently interpolated with trigonometric polynomials by utilizing the computed DFT coefficients. Let $\hat{s} \defeq \ft\{S_f(z)\} = \sum_{l=1}^d \hat{z}^l \pmult \hat{f}^l$ be the DFT of the detection scores $S_f(z)$ evaluated at the sample $z$. The detection scores $s(u,v)$ at the continuous locations $(u,v) \in [0,M) \times [0,N)$ in $z$ are interpolated as,
\begin{equation}
\label{eq:score_interp}
s(u,v) = \frac{1}{MN} \sum_{m=0}^{M-1} \sum_{n=0}^{N-1} \hat{s}(m,n) e^{i2\pi \left(\frac{m}{M}u + \frac{n}{N}v \right)} .
\end{equation}
Here, $i$ denotes the imaginary unit. We aim to find the sub-grid location that corresponds to the maximum score: $(u^*,v^*) = \argmax_{(u,v) \in [0,M) \times [0,N)} s(u,v)$.
The scores $s$ are first evaluated at all grid locations $s(m,n)$ using \eqref{eq:mdcf_detection}. The location of the maximal score $(u^{(0)},v^{(0)}) \in \Omega$ is used as the initial estimate. We then iteratively maximize \eqref{eq:score_interp} using Newton's method, by starting at the location $(u^{(0)},v^{(0)})$. The gradient and Hessian in each iteration are computed by analytically differentiating \eqref{eq:score_interp}. We found that only a few iterations is sufficient for convergence.

We apply the sub-grid interpolation strategy to maximize the classification scores $s_r$ computed at the sample $z_r$. The procedure is applied for each scale level independently. The scale level with the highest maximal detection score is then used to update target location and scale. 

Excluding the feature extraction, the total computational complexity of our tracker sums up to $\mathcal{O}(dSMN\log{MN}+SMNN_\text{Ne}+(d+K^2)dMNN_\text{GS})$. Here, $N_\text{Ne}$ denotes the number of iterations in the sub-grid detection. In our case, the expression is dominated by the last term, which originates from the filter optimization.

\section{Experiments}
Here, we present a comprehensive evaluation of the proposed method. Result are reported on four benchmark datasets: OTB-2013, OTB-2015, ALOV++ and VOT2014.

\subsection{Details and Parameters}
\label{sec:details}
The weight function $w$ is constructed by starting from a quadratic function $w(m,n) = \mu + \eta (m/P)^2 + \eta (n/Q)^2$ with the minimum located at the sample center. Here $P \times Q$ denotes the target size, while $\mu$ and $\eta$ are parameters. The minimum value of w is set to $\mu = 0.1$ and the impact of the regularizer is set to $\eta = 3$. In practice, only a few DFT coefficients in the resulting function have a significant magnitude. We simply remove all DFT coefficients smaller than a threshold to ensure a sparse spectrum $\hat{w}$, containing about 10 non-zero coefficients. Figure~\ref{fig:periodic} visualizes the resulting weight function $w$ used in the optimization.

Similar to recent DCF based trackers \cite{DanelljanBMVC14,HenriquesPAMI15,Li2014}, we also employ HOG features, using a cell size of $4 \times 4$ pixels. Samples are represented by a square $M \times N$ grid of cells (\ie $M = N$), such that the corresponding image area is proportional to the area of the target bounding box. We set the image region area of the samples to $4^2$ times the target area and set the initial scale to ensure a maximum sample size of $M=50$ cells. Samples are multiplied by a Hann window \cite{MOSSE2010}.
We set the label function $y_t$ to a sampled Gaussian with a standard deviation proportional to the target size \cite{DanelljanBMVC14,Henriques12d}. The learning rate is set to $\gamma = 0.025$ and we use $N_\text{GS} = 4$ Gauss-Seidel iterations. All parameters remain fixed for all videos and datasets. Our Matlab implementation\footnote{Available at \url{http://www.cvl.isy.liu.se/research/objrec/visualtracking/regvistrack/index.html}.} runs at 5 frames per second on a standard desktop computer.

\subsection{Baseline Comparison}
Here, we evaluate the impact of the proposed spatial regularization component and compare it with the standard DCF formulation. First, we investigate the consequence of simply replacing the proposed regularizer with the standard DCF regularization in \eqref{eq:mdcf_cost}, without altering any parameters. This corresponds to using uniform regularization weights $w(m,n) = \sqrt{\lambda}$, in our framework. We set $\lambda = 0.01$ following \cite{DanelljanBMVC14,DanelljanCVPR14,Henriques12d}. For a fair comparison, we also evaluate both our and the standard regularization using a smaller sample size relative to the target, by setting the size as in \cite{DanelljanBMVC14,DanelljanCVPR14,Henriques12d}.

Table~\ref{tab:baseline_reg} shows the mean overlap precision (OP) for the four methods on the OTB-2013 dataset. The OP is computed as the fraction of frames in the sequence where the intersection-over-union overlap with the ground truth exceeds a threshold of $0.5$ (PASCAL criterion).
The standard DCF benefits from using smaller samples to avoid corrupting the positive training samples with background information. On the other hand, the proposed spatial regularization enables an expansion of the image region used for training the filter, without corrupting the target model. This leads to a more discriminative model, resulting in a gain of $7.0 \%$ in mean OP compared to the standard DCF formulation.

Additionally, we compare our method with Correlation Filters with Limited Boundaries (CFLB) \cite{GaloogahiCVPR2015}. For a fair comparison, we use the same settings as in \cite{GaloogahiCVPR2015} for our approach: single grayscale channel, no scale estimation, no sub-grid detection and the same sample size. On the OTB-2013, the CFLB achieves a mean OP of $48.6 \%$. Whereas the mentioned baseline version of our tracker obtains a mean OP of $54.3 \%$, outperforming \cite{GaloogahiCVPR2015} by $5.7 \%$.

\begin{table}[!t]
	\centering
	\resizebox{0.48\textwidth}{!}{\begin{tabular}{lcccc}
\toprule
&\multicolumn{2}{c}{\hspace{-5mm}\textbf{Conventional sample size}}&\multicolumn{2}{c}{\textbf{Expanded sample size}}\\
Regularization&Standard&Ours&\hspace{3mm}Standard&Ours\\\midrule
Mean OP ($\%$)&71.1&72.2&\hspace{3mm}50.1&\textbf{\textcolor{red}{78.1}}\\\bottomrule
\end{tabular}
}\vspace{1mm}
	\caption{
		A comparison of tracking performance on OTB-2013 when using the standard regularization \eqref{eq:mdcf_cost} and the proposed spatial regularization \eqref{eq:ours_cost_spatial}, in our tracking framework. The comparison is performed both with a conventional sample size (used in existing DCF based trackers) and our expanded sample size.
		}
	\label{tab:baseline_reg}
\end{table}

\vspace{1mm}
\subsection{OTB-2013 Dataset}

\begin{table}
	\centering
	\resizebox{0.48\textwidth}{!}{\begin{tabular}{l@{~}c@{~~}c@{~~}c@{~~}c@{~~}c@{~~}c@{~~}c@{~~}c@{~~}c@{~~}c}
\toprule
&LSHT&ASLA&Struck&ACT&TGPR&KCF&DSST&SAMF&MEEM&\textbf{SRDCF}\\\midrule
OTB-2013&47.0&56.4&58.8&52.6&62.6&62.3&67&69.7&\textit{\textcolor{blue}{70.1}}&\textbf{\textcolor{red}{78.1}}\\\midrule
OTB-2015&40.0&49.0&52.9&49.6&54&54.9&60.6&\textit{\textcolor{blue}{64.7}}&63.4&\textbf{\textcolor{red}{72.9}}\\\bottomrule
\end{tabular}
}\vspace{1mm}
	\caption{A comparison with state-of-the-art trackers on the OTB-2013 and OTB-2015 datasets using mean overlap precision (in percent). The best two results for each dataset are shown in red and blue fonts respectively. Our SRDCF achieves a gain of $8.0 \%$ and $8.2 \%$ on OTB-2013 and OTB-2015 respectively compared to the second best tracker on each dataset.
	}\vspace{-2mm}
	\label{tab:OTB_comparison}
\end{table}

We provide a comparison of our tracker with 24 state-of-the-art methods from the literature: MIL \cite{Babenko09b}, IVT \cite{Ross08d}, CT \cite{Zhang12c}, TLD \cite{Mikolajczyk10d}, DFT \cite{Laura12d}, EDFT \cite{Felsberg13c}, ASLA \cite{Jia12d}, L1APG \cite{Bao12d}, CSK \cite{Henriques12d}, SCM \cite{Zhong12g}, LOT \cite{Oron12b}, CPF \cite{Perez02b}, CXT \cite{Dinh11}, Frag \cite{Adam6c}, Struck \cite{Torr11b}, LSHT \cite{Shengfeng13b}, LSST \cite{Wang13d}, ACT \cite{DanelljanCVPR14}, KCF \cite{HenriquesPAMI15}, CFLB \cite{GaloogahiCVPR2015}, DSST \cite{DanelljanBMVC14}, SAMF \cite{Li2014}, TGPR \cite{TGPR2014} and MEEM \cite{MEEM2014}.

\subsubsection{State-of-the-art Comparison}
Table~\ref{tab:OTB_comparison} shows a comparison with state-of-the-art methods on the OTB-2013 dataset, using mean overlap precision (OP) over all 50 videos. Only the results for the top 10 trackers are reported.
The MEEM tracker, based on an online SVM, provides the second best results with a mean OP of $70.1\%$. The best result on this dataset is obtained by our tracker with a mean OP of $78.1\%$, leading to a significant gain of $8.0 \%$ compared to MEEM.

\begin{figure}[!t]
	\centering\vspace{-3.5mm}
	\newcommand{\wid}{0.24\textwidth}
	\subfloat[OTB-2013\label{fig:sota_ope}]{\includegraphics[width = \wid]{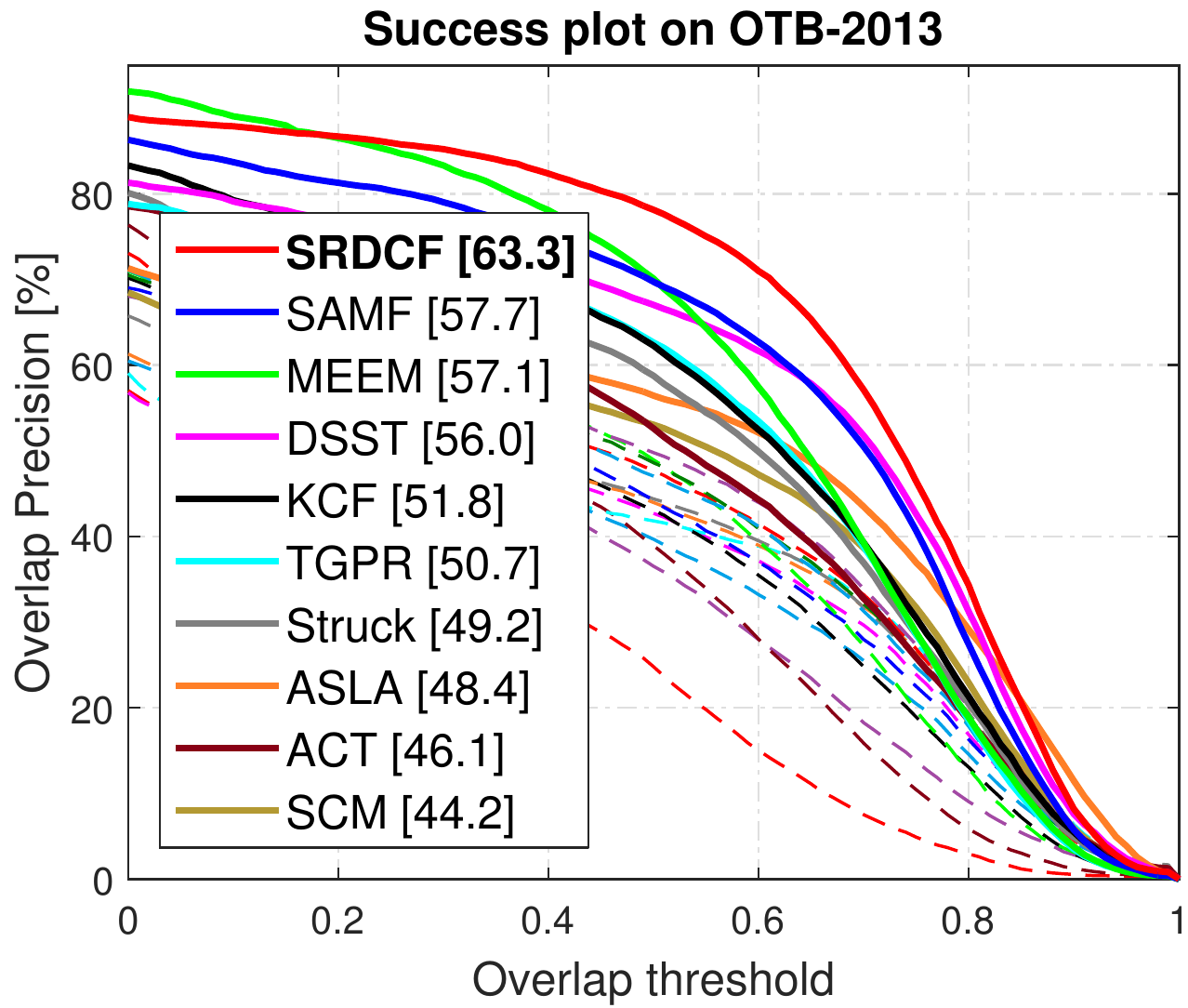}}
	\subfloat[OTB-2015\label{fig:sota_ope_100}]{\includegraphics[width = \wid]{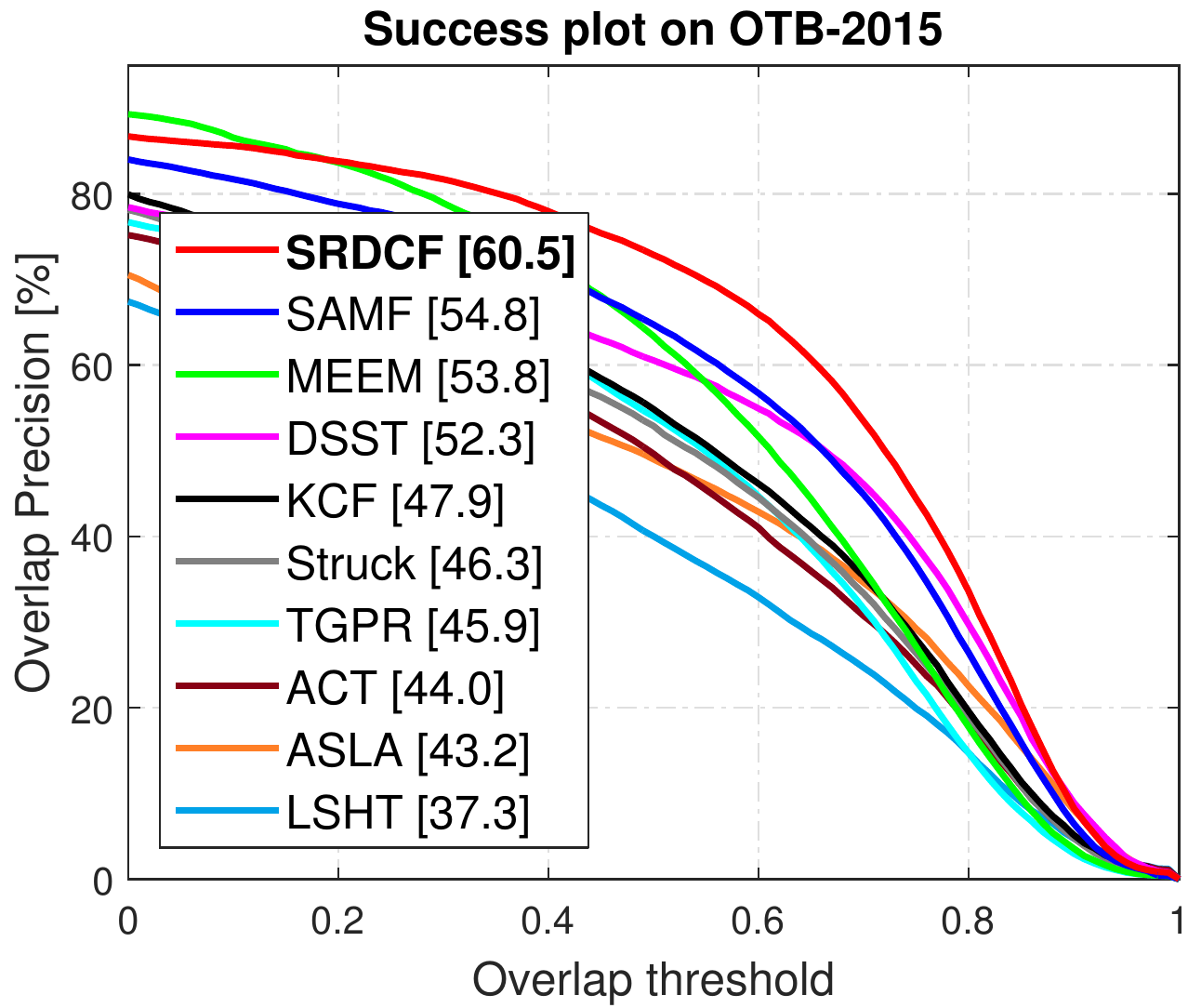}}
	\caption{Success plots showing a comparison with state-of-the-art methods on OTB-2013 (a) and OTB-2015 (b). For clarity, only the top 10 trackers are displayed. Our SRDCF achieves a gain of $5.6 \%$ and $5.7 \%$ on OTB-2013 and OTB-2015 respectively, compared to the second best methods.}\vspace{-2mm}
	\label{fig:OTB_ope}
\end{figure}

Figure~\ref{fig:sota_ope} shows the success plot over all the 50 videos in OTB-2013. The success plot shows the mean overlap precision (OP), plotted over the range of intersection-over-union thresholds. The trackers are ranked using the \emph{area under the curve} (AUC), displayed in the legend.
Among previous DCF based trackers, DSST and SAMF provides the best performance, with an AUC score of $56.0\%$ and $57.7\%$. Our approach obtains an AUC score of $63.3\%$ and significantly outperforms the best existing tracker (SAMF) by $5.6\%$.  

\begin{figure}[!t]
	\centering
	\newcommand{\wid}{0.24\textwidth}
	\includegraphics[width = \wid]{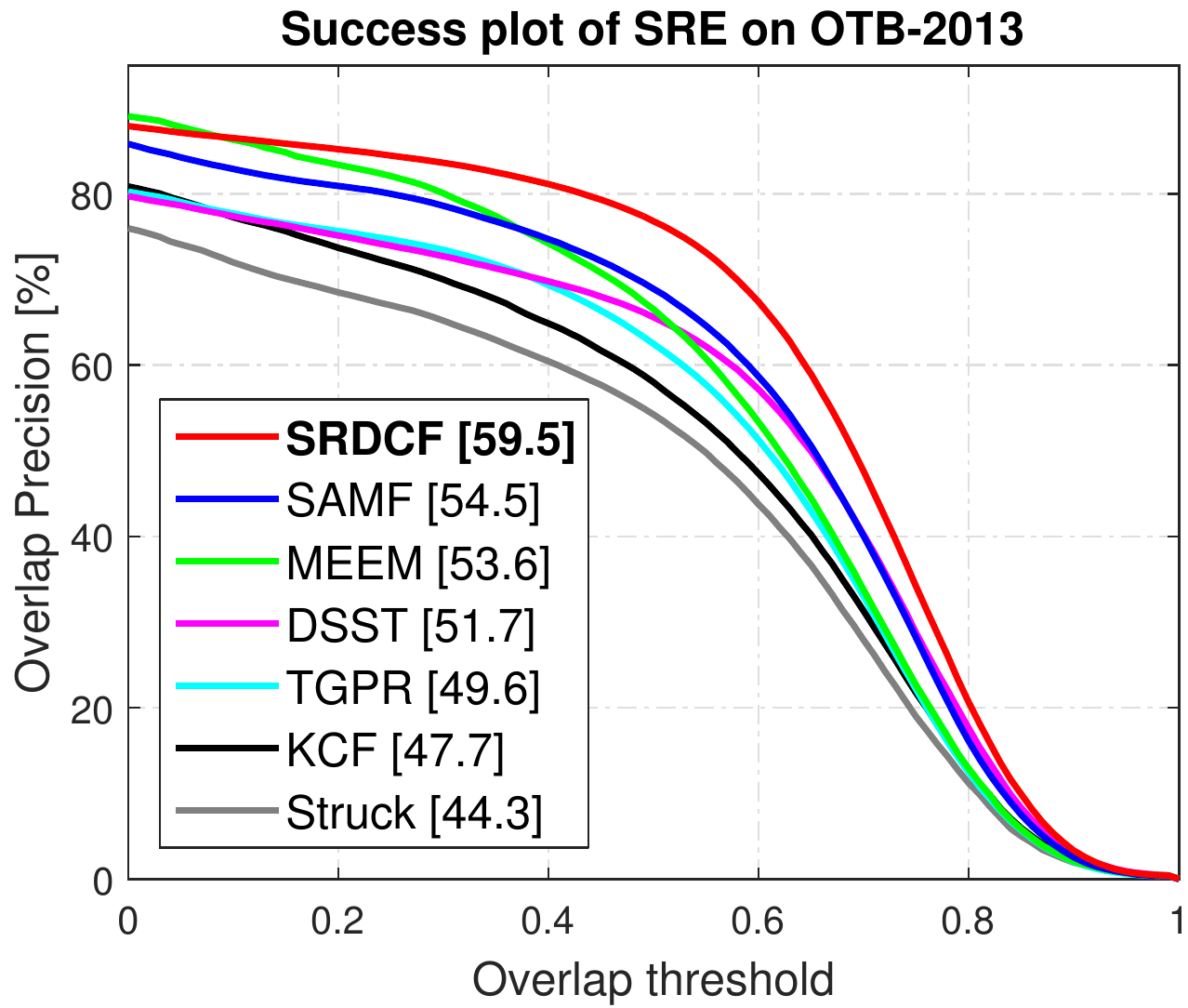}%
	\includegraphics[width = \wid]{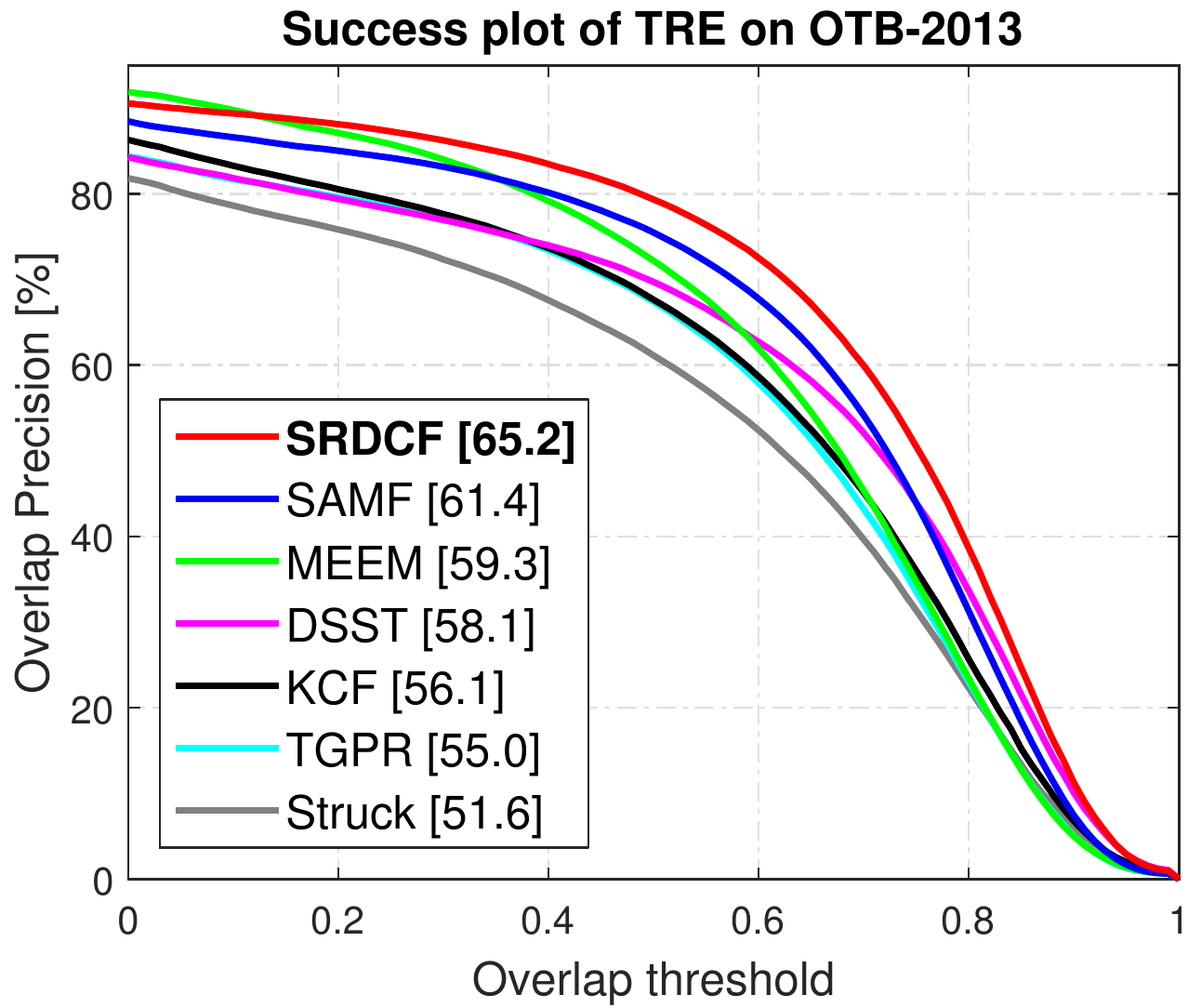}\vspace{-1mm}
	\caption{Comparison with respect to robustness to initialization on OTB-2013. We show success plots for both the spatial (SRE) and temporal (TRE) robustness. Our approach clearly demonstrates robustness in both scenarios.}\vspace{-2mm}
	\label{fig:sota_sretre}
\end{figure}

\subsubsection{Robustness to Initialization}
Visual tracking methods are known to be sensitive to initialization. We evaluate the robustness of our tracker by  following the protocol proposed in \cite{Wu13g}. Two different types of initialization criteria, namely: temporal robustness (TRE) and spatial robustness (SRE), are evaluated. The SRE corresponds to tracker initialization at different positions close to the ground-truth in the first frame. The procedure is repeated with 12 different initializations for each video in the dataset. The TRE criteria evaluates the tracker by initializations at 20 different frames, with the ground-truth.

\begin{figure}[!t]
	\centering
	\newcommand{\wid}{0.12\textwidth}
	\includegraphics*[trim = 130 50 70 0,width = \wid]{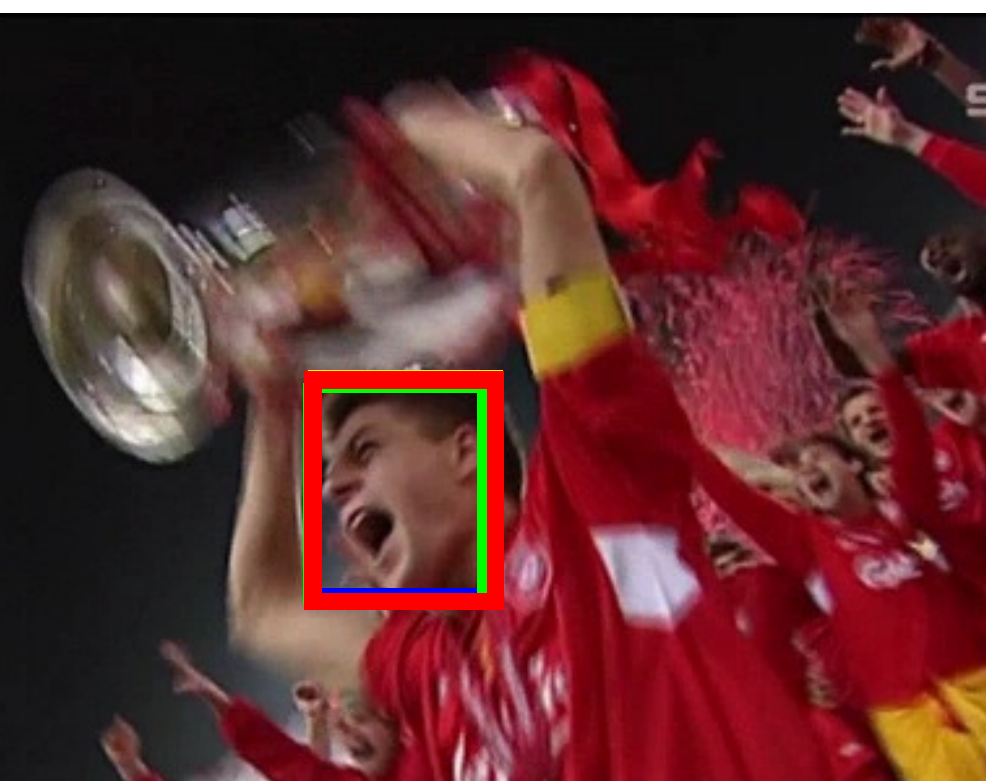}%
	\includegraphics*[trim = 130 50 70 0,width = \wid]{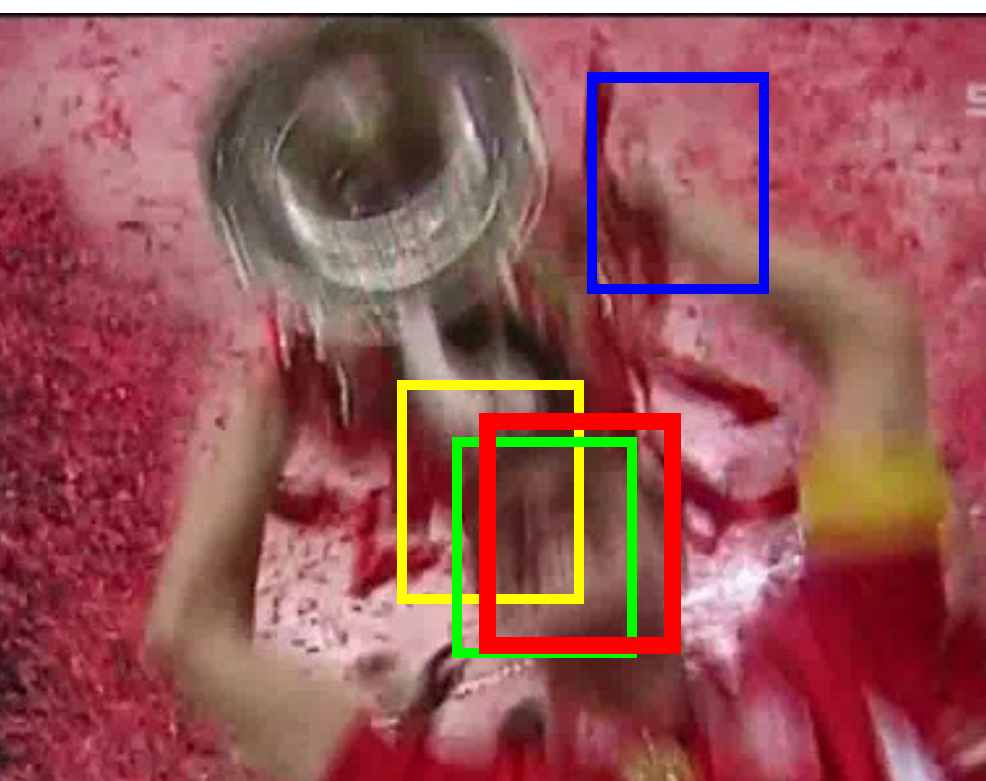}%
	\includegraphics*[trim = 70 50 130 0,width = \wid]{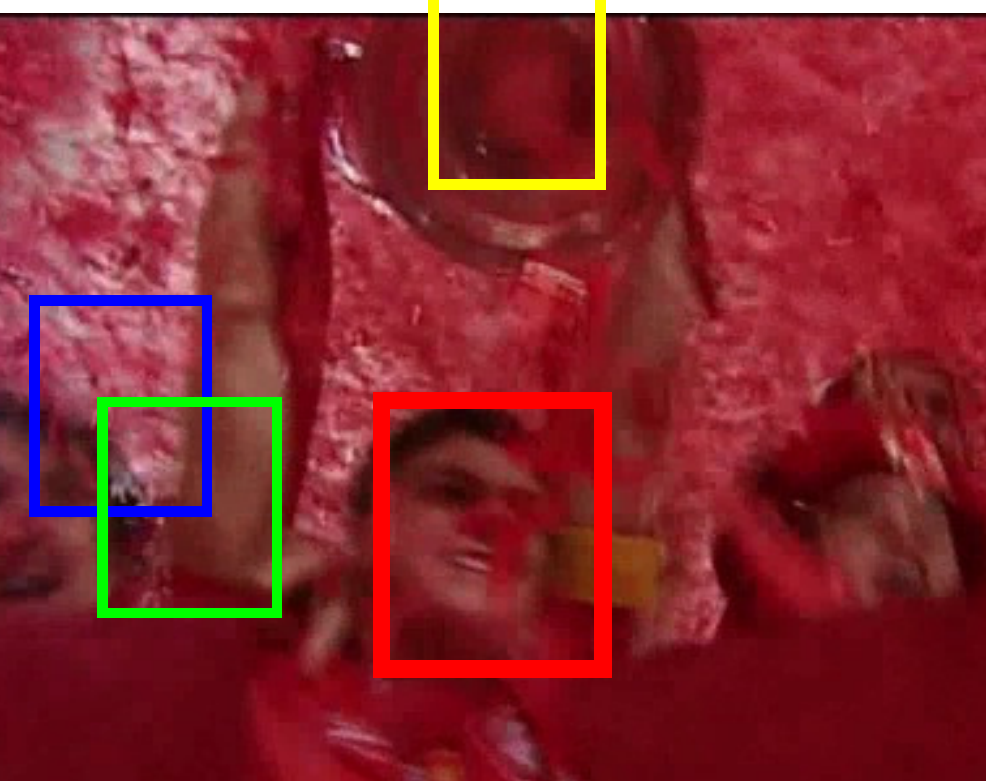}%
	\includegraphics*[trim = 70 50 130 0,width = \wid]{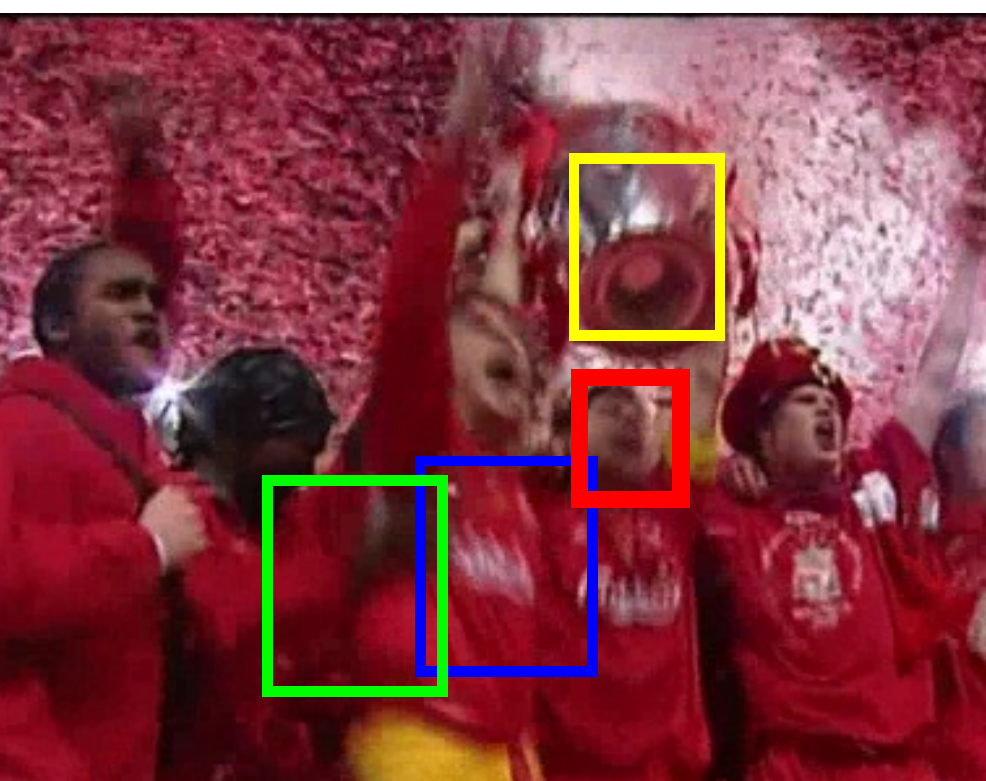}
	\includegraphics*[trim = 50 50 5 180,width = \wid]{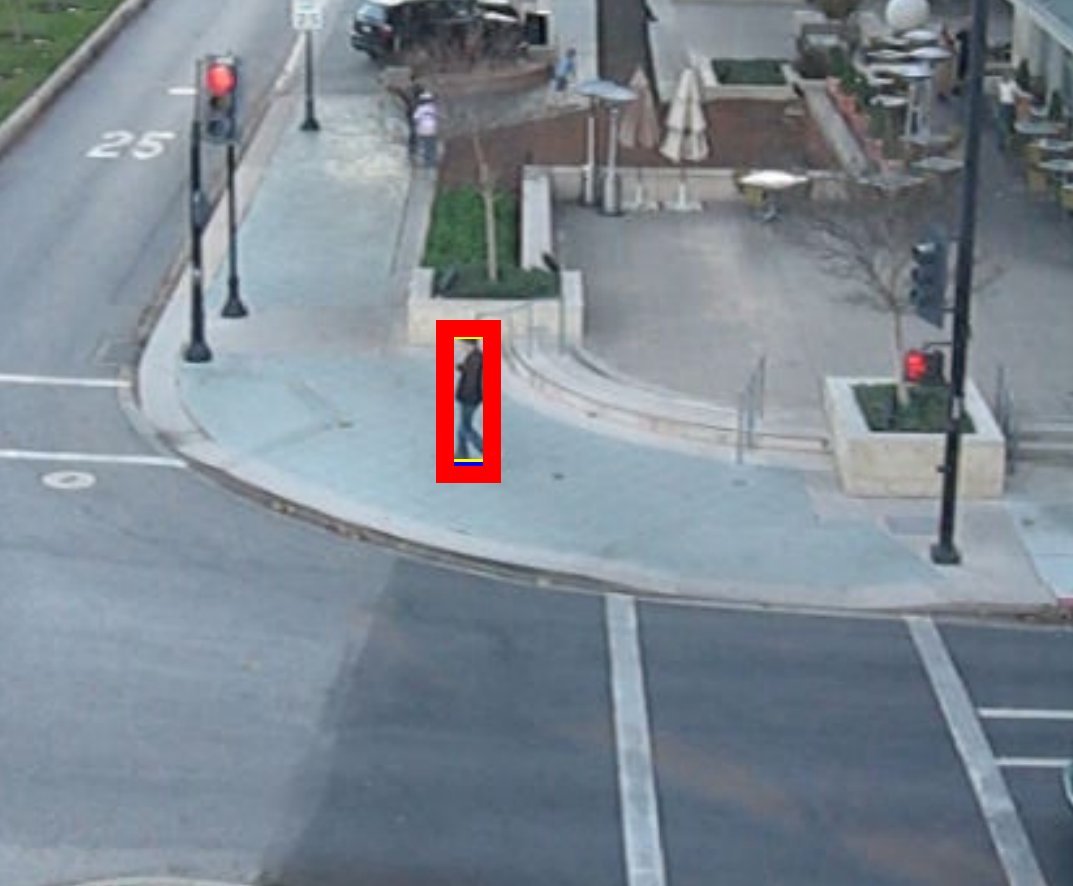}%
	\includegraphics*[trim = 50 0 5 230,width = \wid]{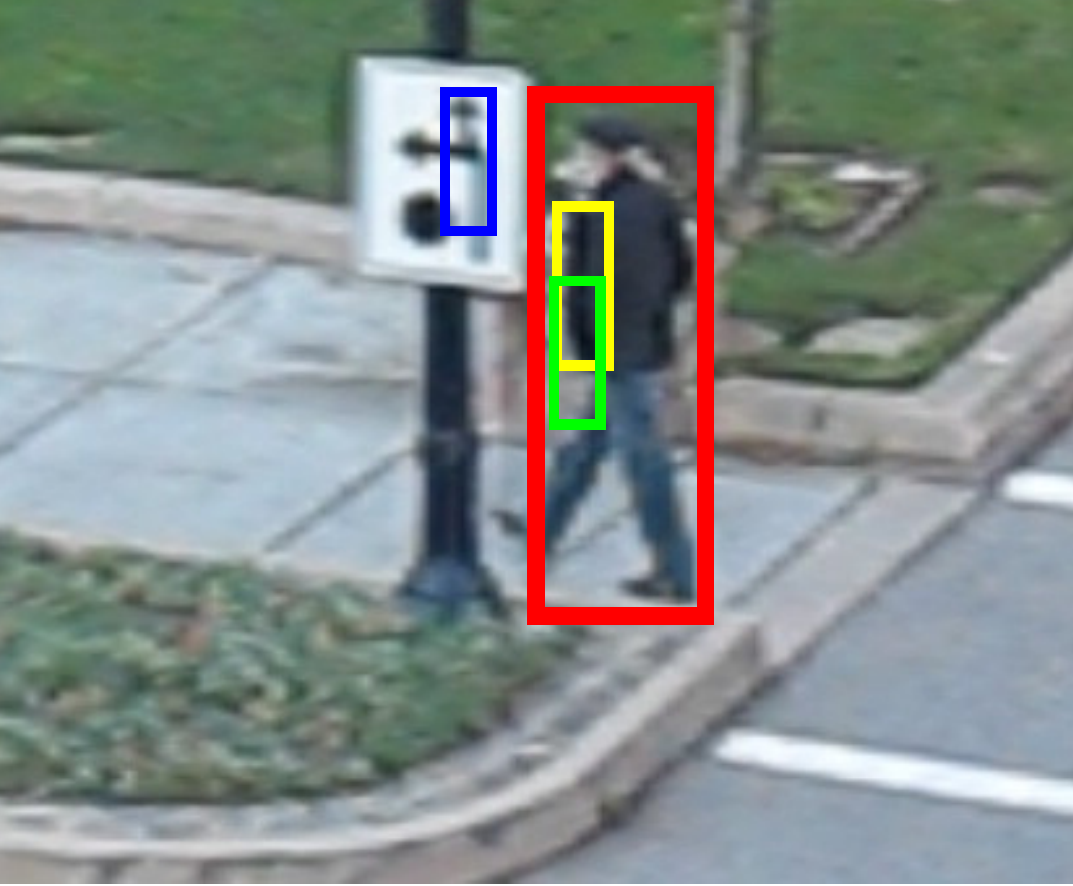}%
	\includegraphics*[trim = 50 0 5 230,width = \wid]{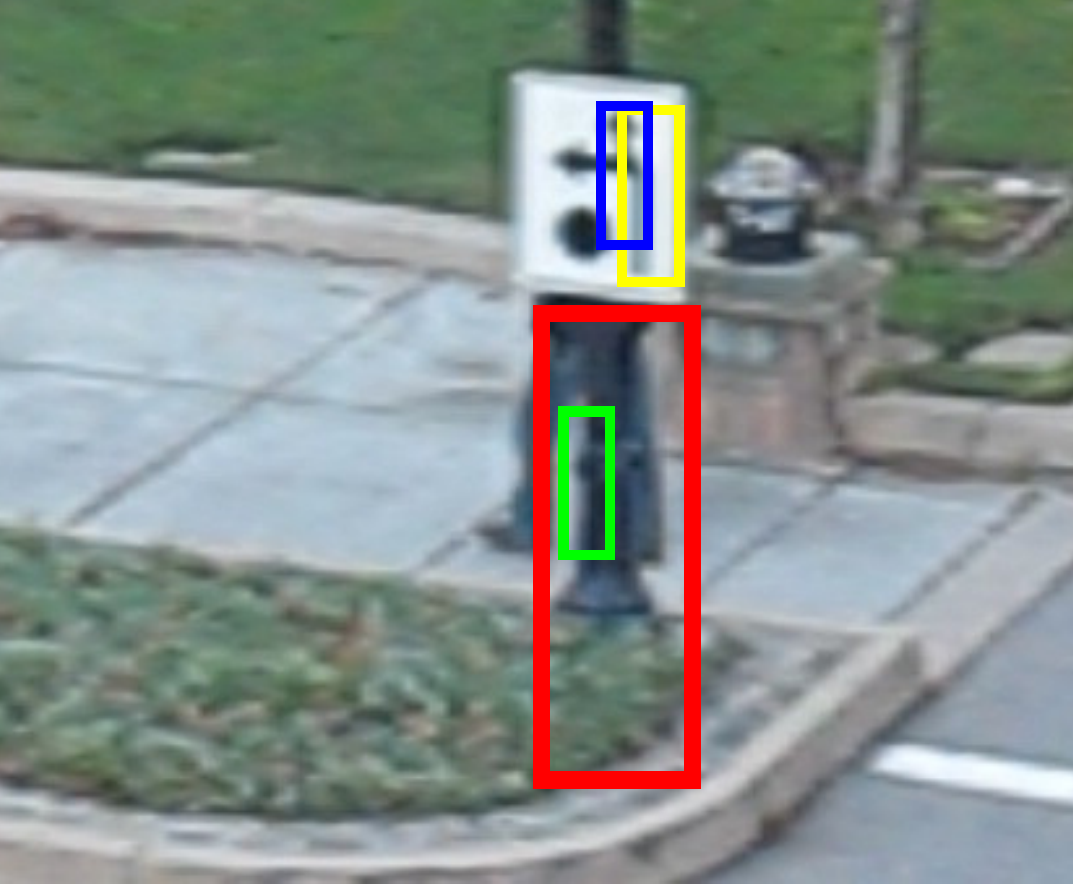}%
	\includegraphics*[trim = 50 0 5 230,width = \wid]{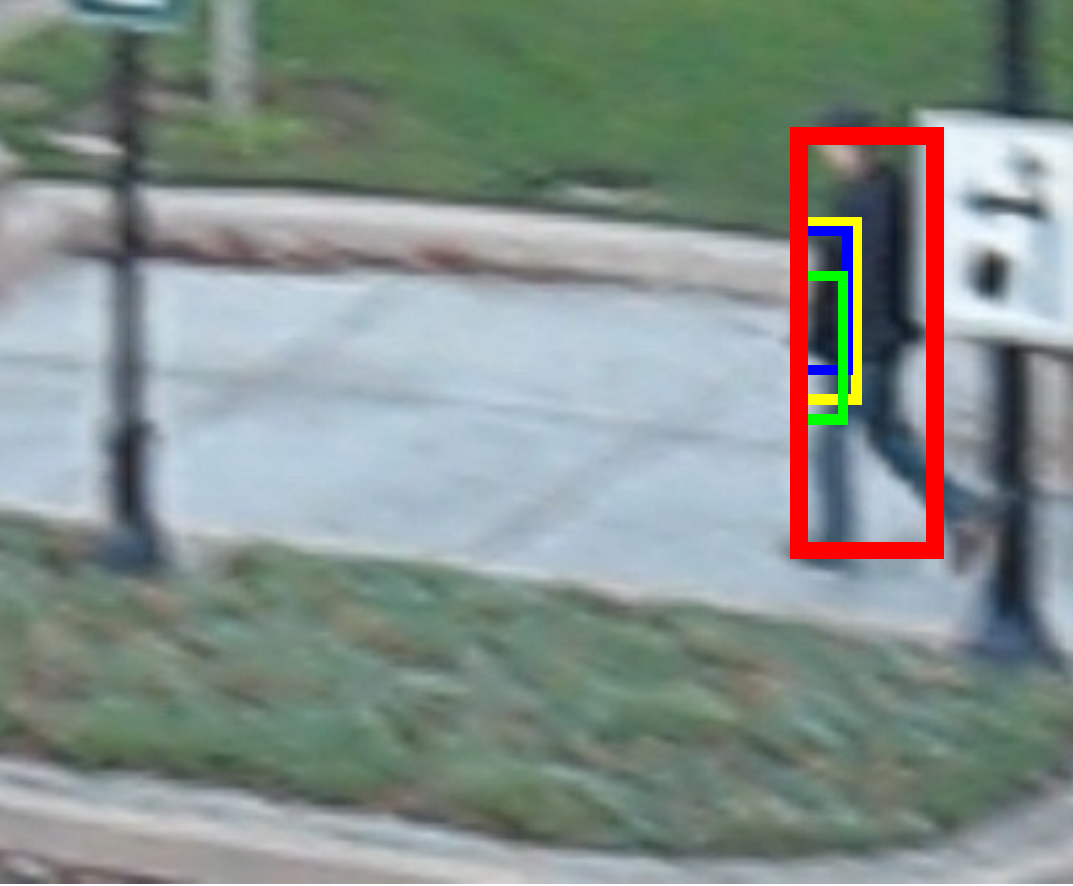}
	\includegraphics*[trim = 50 180 230 30,width = \wid]{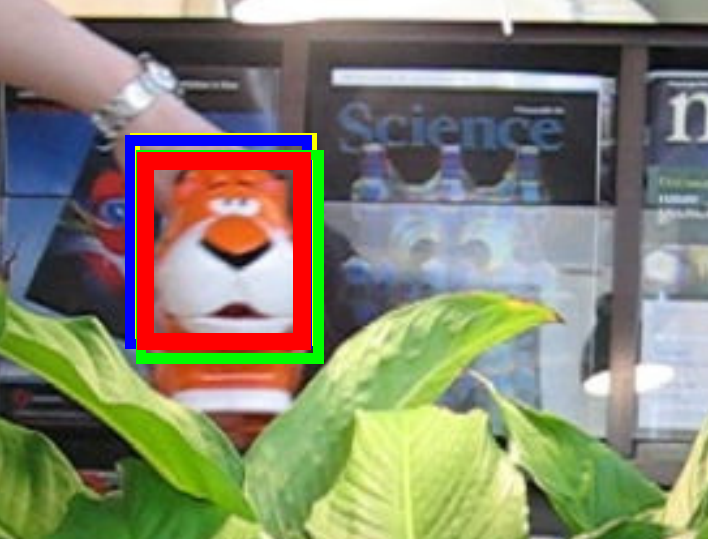}%
	\includegraphics*[trim = 100 180 180 30,width = \wid]{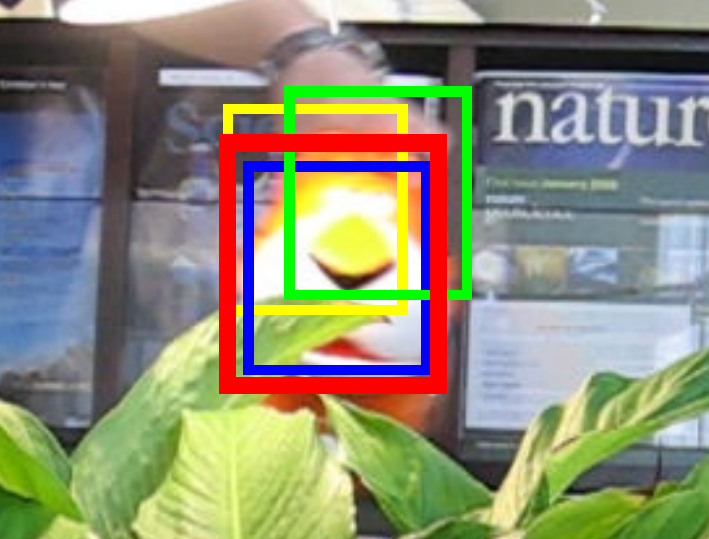}%
	\includegraphics*[trim = 100 180 180 30,width = \wid]{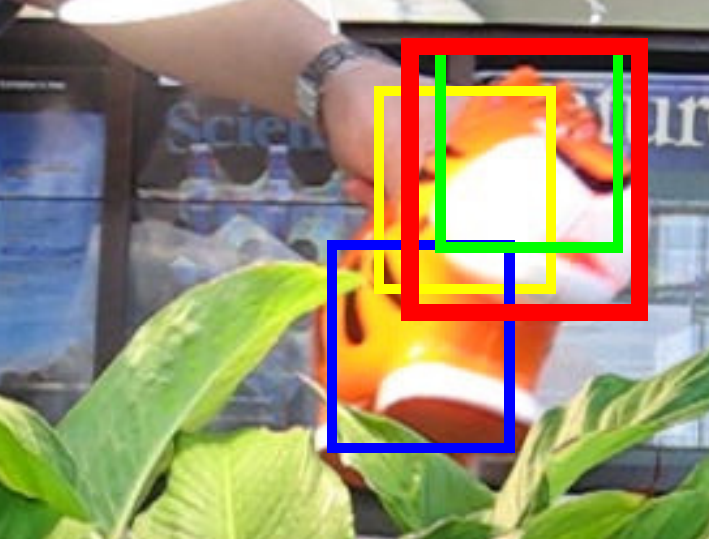}%
	\includegraphics*[trim = 50 180 230 30,width = \wid]{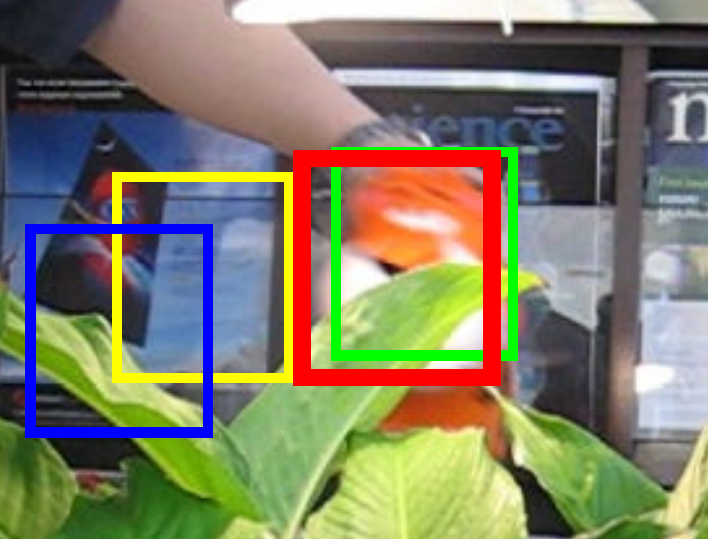}
	\includegraphics*[trim = 2 2 2 5,width = 0.40\textwidth]{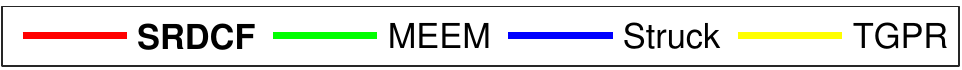}\vspace{-1mm}
	\caption{Qualitative comparison of our approach with state-of-the-art trackers on the \emph{Soccer}, \emph{Human6} and \emph{Tiger2} videos. Our approach provides consistent results in challenging scenarios, such as occlusions, fast motion, background clutter and target rotations.}\vspace{-2mm}
	\label{fig:qualitative}
\end{figure}

Figure~\ref{fig:sota_sretre} shows the success plots for TRE and SRE on the OTB-2013 dataset with 50 videos. We include all the top 7 trackers in figure~\ref{fig:sota_ope} for this experiment. Among the existing methods, SAMF and MEEM provide the best results. Our SRDCF achieves a consistent gain in performance over these trackers on both robustness evaluations. 

\begin{figure*}[!t]
	\centering
	\newcommand{\wid}{0.25\textwidth}
	\newcommand{\name}{figures/sota_OPE}
	\newcommand{\eval}{OPE}
	\includegraphics[width=\wid]{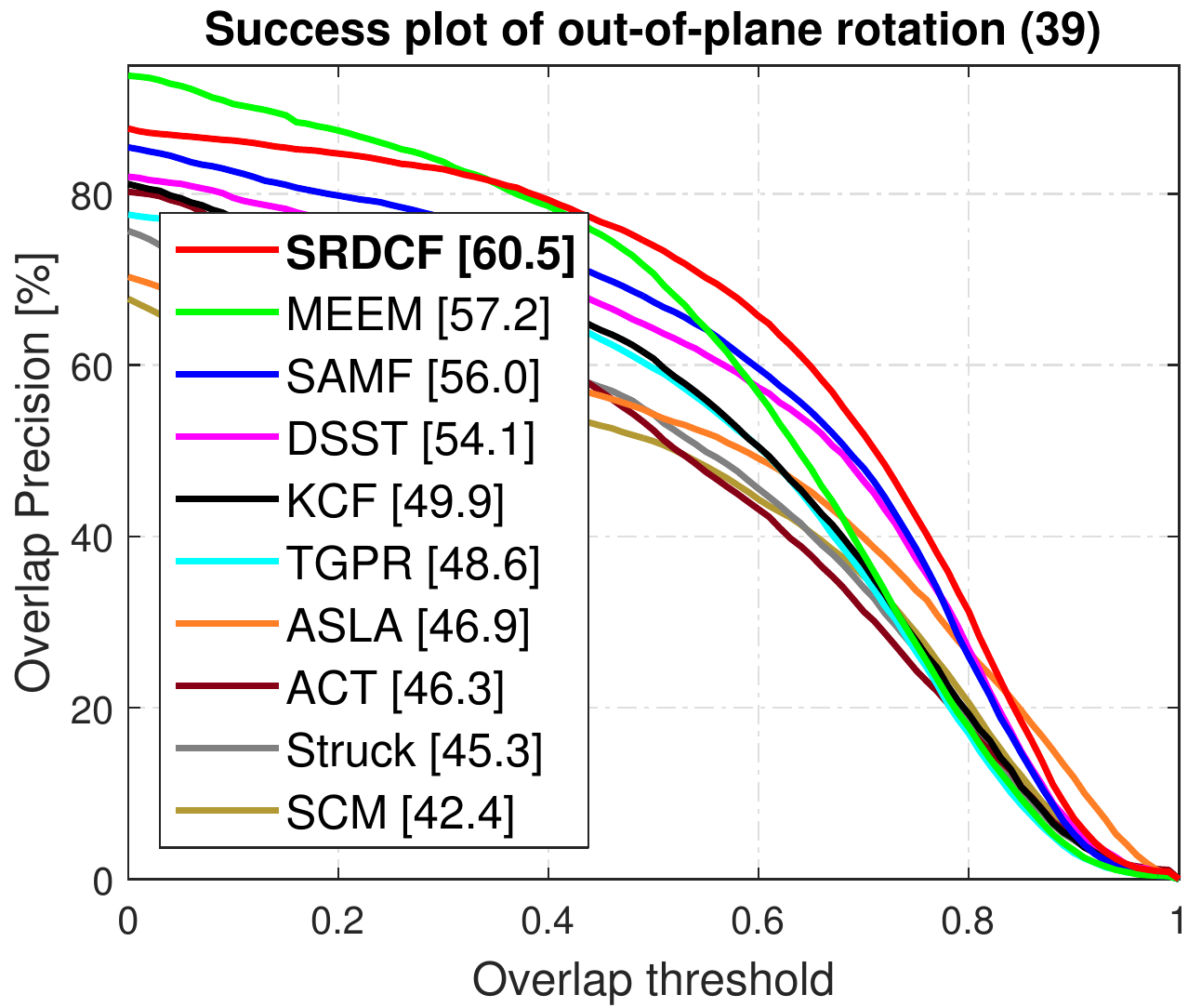}%
	\includegraphics[width=\wid]{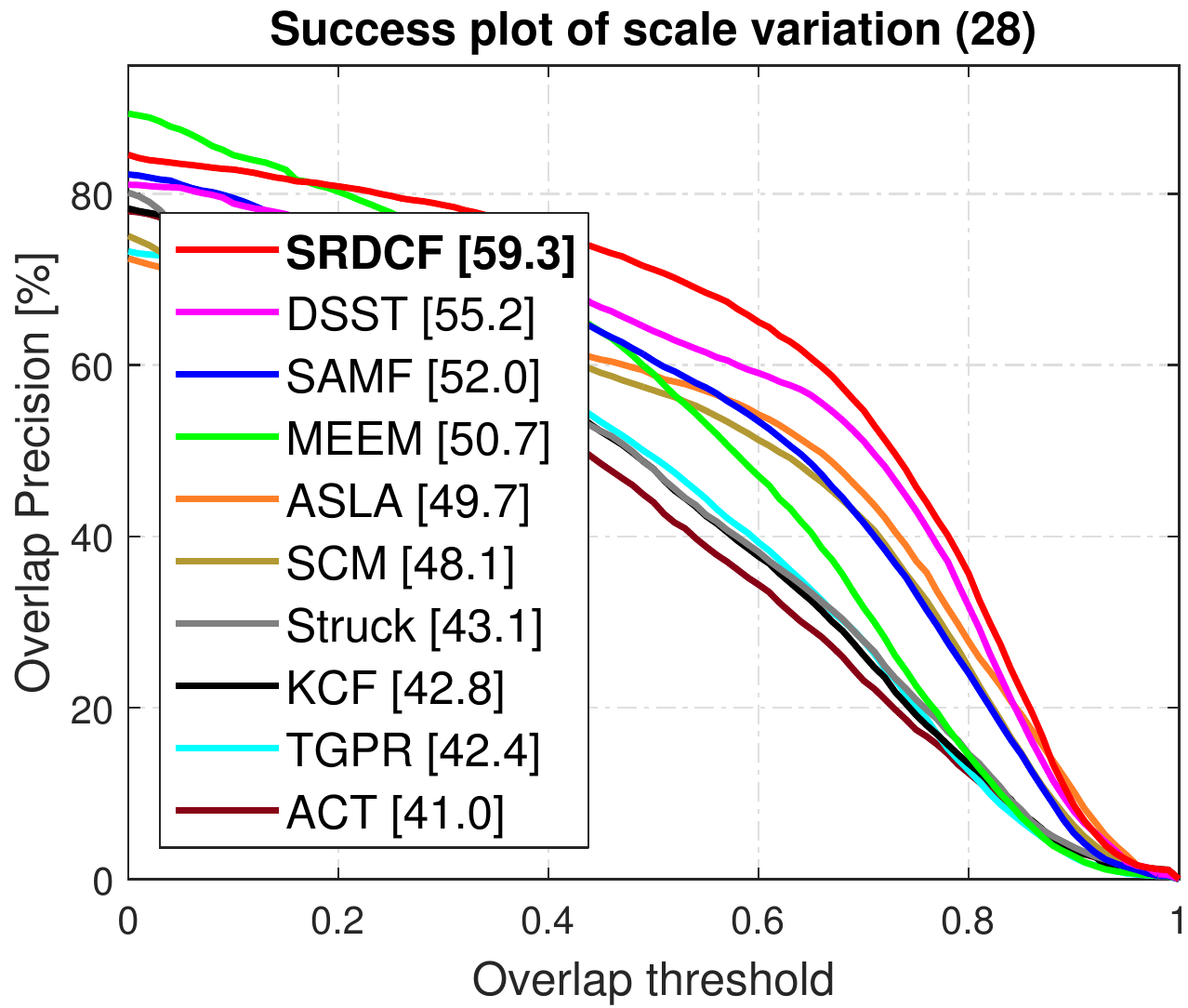}%
	\includegraphics[width=\wid]{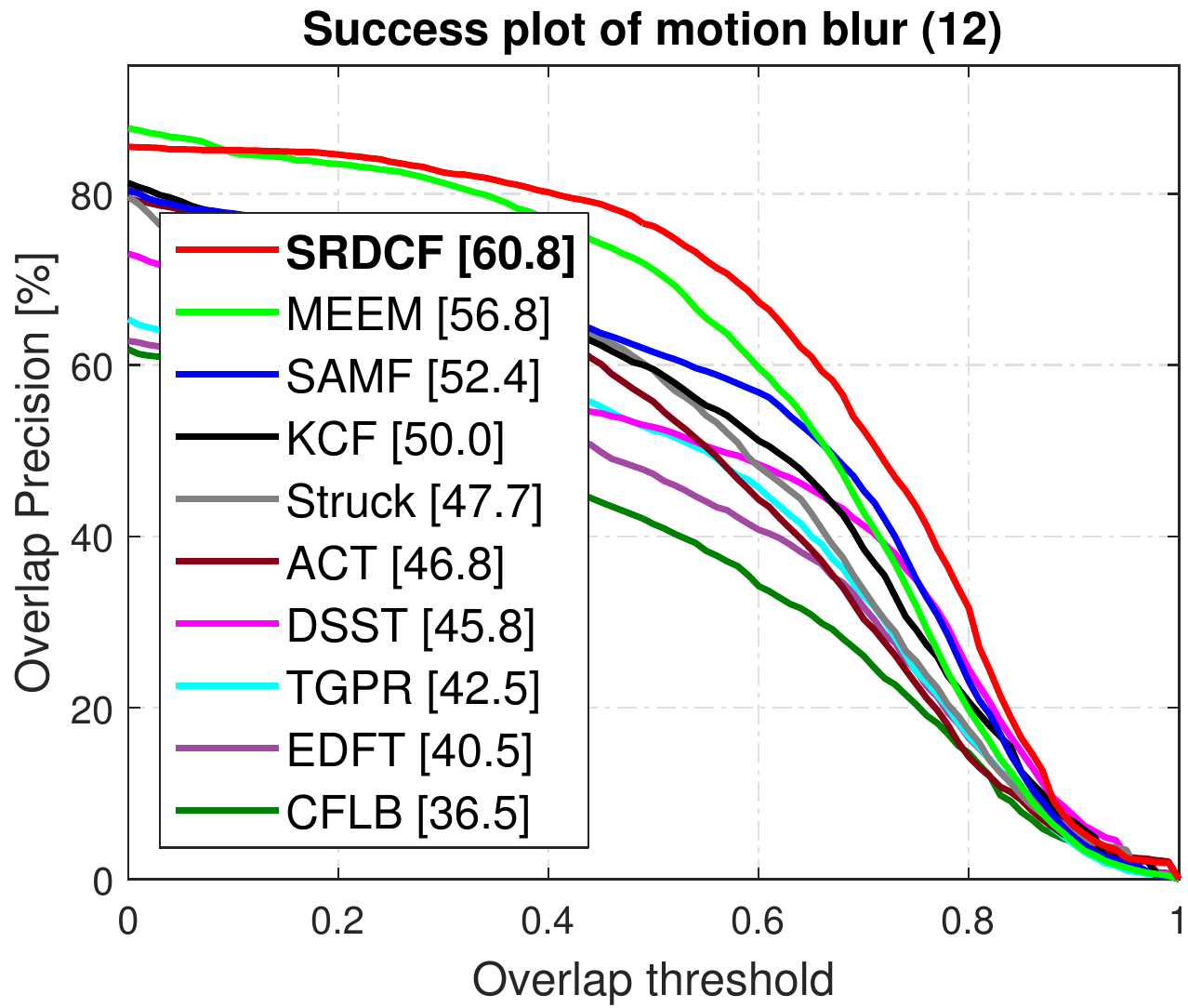}%
	\includegraphics[width=\wid]{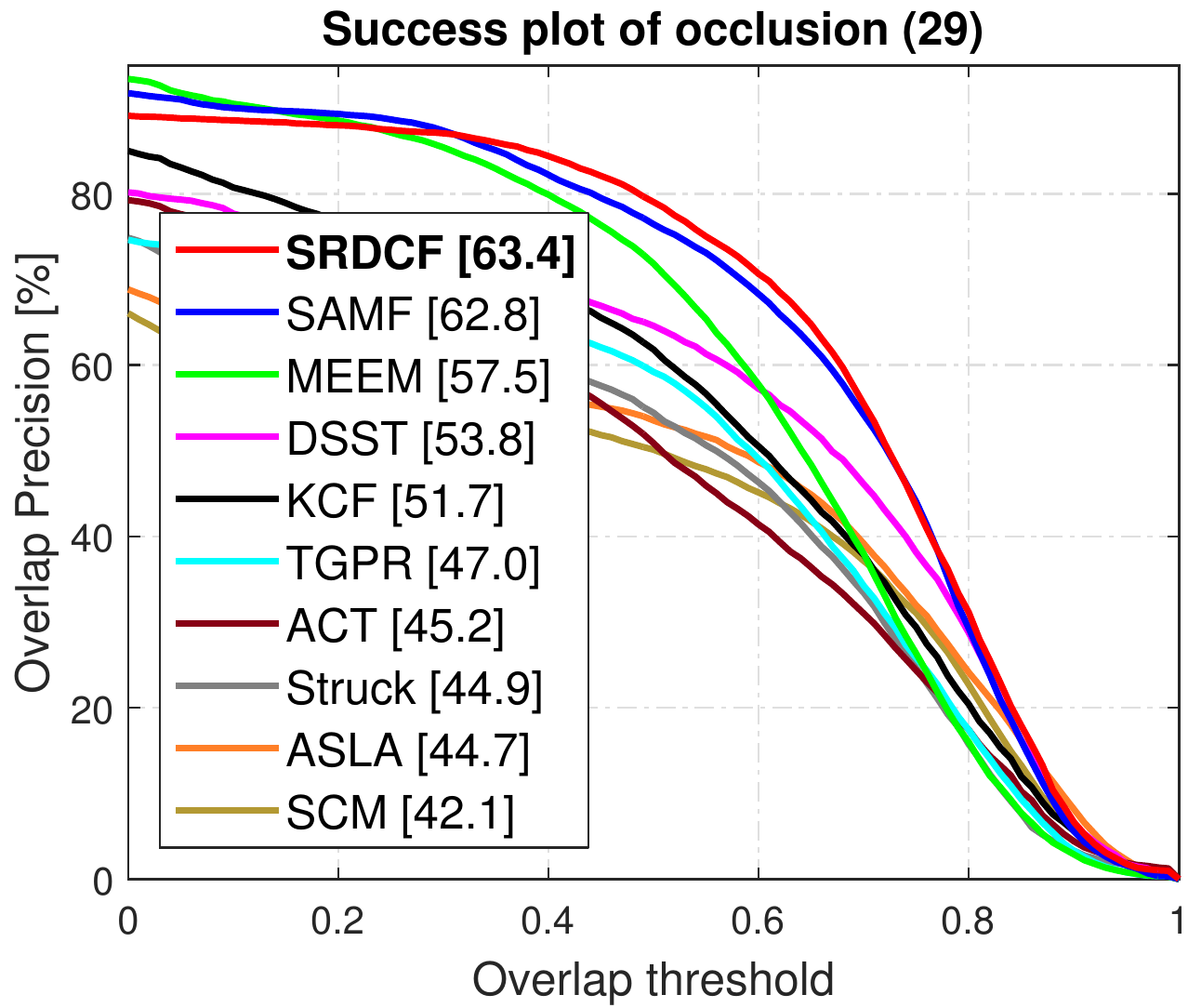}\vspace{-0.5mm}
	\caption{Attribute-based analysis of our approach on the OTB-2013 dataset with 50 videos. Success plots are shown for four attributes. Each plot title includes the number of videos associated with the respective attribute. Only the top 10 trackers for each attribute are displayed for clarity. Our approach demonstrates superior performance compared to existing trackers in these scenarios.}\vspace{-3mm}
	\label{fig:attribute}
\end{figure*}

\subsubsection{Attribute Based Comparison}
We perform an attribute based analysis of our approach on the OTB-2013 dataset. All the 50 videos in OTB-2013 are annotated with 11 different attributes, namely: illumination variation, scale variation, occlusion,  deformation,  motion  blur, fast motion, in-plane rotation, out-of-plane rotation, out-of-view, background clutter and low resolution. Our approach outperforms existing trackers on 10 attributes.

Figure~\ref{fig:attribute} shows example success plots of four different attributes. Only the top 10 trackers in each plot are displayed for clarity. In case of out-of-plane rotations, (MEEM) achieves an AUC score of $57.2\%$. Our tracker provides a gain of $3.3\%$ compared to MEEM. Among the existing methods, the two DCF based trackers DSST and SAMF provide the best results in case of scale variation. Both these trackers are designed to handle scale variations. Our approach achieves a significant gain of $4.1\%$ over DSST. Note that the standard DCF trackers struggle in the cases of  motion blur and fast motion due to the restricted search area. This is caused by the induced boundary effects in the detection samples of the standard DCF trackers. Our approach significantly improves the performance compared to the standard DCF based trackers in these cases. 
Figure~\ref{fig:qualitative} shows a qualitative comparison of our approach with existing methods on challenging example videos. 
Despite no explicit occlusion handling component, our tracker performs favorably in cases with occlusion.

\subsection{OTB-2015 Dataset}
We provide a comparison of our approach on the recently introduced OTB-2015. The dataset extends OTB-2013 and contains 100 videos. Table~\ref{tab:OTB_comparison} shows the comparison with the top 10 methods, using mean overlap precision (OP) over all 100 videos. Among the existing methods, SAMF and MEEM provide the best results with mean OP of $64.7\%$ and $63.4\%$ respectively. Our tracker outperforms the best existing tracker by $8.2\%$ in mean OP.

Figure~\ref{fig:sota_ope_100} shows the success plot over all the 100 videos. Among the standard DCF trackers, SAMF provides the best results with an AUC score of $54.8\%$. The MEEM tracker achieves an AUC score of $53.8\%$. Our tracker obtains an AUC score of $60.5\%$, outperforming SAMF by $5.7\%$.

\begin{figure}[!t]
	\centering
	\includegraphics[width = 0.47\textwidth]{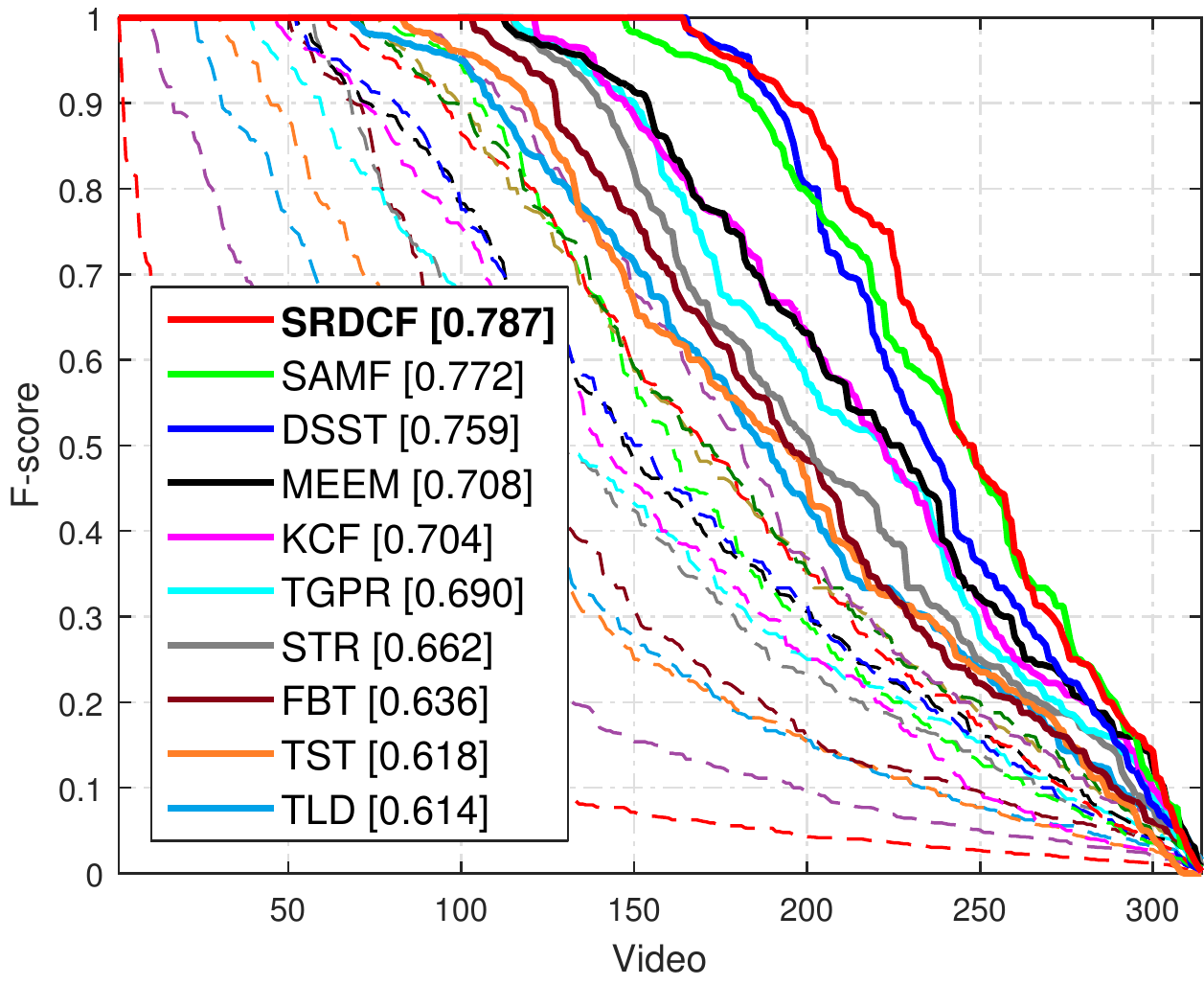}\vspace{0mm}
	\caption{Survival curves comparing our approach with 24 trackers on ALOV++. The mean F-scores for the top 10 trackers are shown in the legend. Our approach achieves the best overall results.}\vspace{-3mm}
	\label{fig:alov}
\end{figure}

\subsection{ALOV++ Dataset}
We also perform experiments on the ALOV++ dataset \cite{ALOV}, containing 314 videos with 89364 frames in total. The evaluation protocol employs survival curves based on F-score, where a higher F-score indicates better performance.
The survival curve is constructed by plotting the sorted F-scores of all 314 videos.
We refer to \cite{ALOV} for details.

Our approach is compared with the 19 trackers evaluated in \cite{ALOV}. We also add the top 5 methods from our OTB comparison. Figure~\ref{fig:alov} shows the survival curves and the average F-scores of the trackers. MEEM obtains a mean F-score of $0.708$. Our approach obtains the best overall performance compared to 24 trackers with a mean F-score of $0.787$.

\subsection{VOT2014 Dataset}
Finally, we present results on VOT2014 \cite{VOT2014}. Our approach is compared with the 38 participating trackers in the challenge. We also add MEEM in the comparison. In VOT2014, the trackers are evaluated both in terms of accuracy and robustness. The accuracy score is based on the overlap with ground truth, while the robustness is determined by the failure rate.
The trackers are restarted at each failure. The final rank is based on the accuracy and robustness in each video. We refer to \cite{VOT2014} for details.

Table~\ref{tab:VOT_comparison} shows the final ranking scores over all the videos in VOT2014. Among the existing methods, the DSST approach provides the best results. Our tracker achieves the top final rank of $8.26$, outperforming DSST and SAMF. 

\begin{table}[!t]
	\centering
	\resizebox{0.47\textwidth}{!}{\newcommand{\first}[1]{\textbf{\textcolor{red}{#1}}}%
\newcommand{\second}[1]{\textit{\textcolor{blue}{#1}}}%
\begin{tabular}{lccccc}
\toprule
& Overlap & Failures & Acc.\ Rank & Rob.\ Rank & Final Rank \\\midrule
\textbf{SRDCF} & 0.63 & 15.90 & 6.43 & 10.08 & \first{8.26}\\
DSST & 0.64	& 16.90 & 5.99 & 11.17 & \second{8.58} \\
SAMF & 0.64	& 19.23 & 5.87 & 14.14 & 10.00 \\
%PLT14 & 0.55 & 3.41 & 14.95 & 5.67 & 10.31 \\
%KCF & 0.66 & 19.79 & 5.56 & 16.67 & 11.11 \\
%PLT13 & 18.94 & 3.83 & 11.39 \\
%DGT & 11.62 & 11.42 & 11.52 \\
%MCT & 17.21 & 11.39 & 14.30 \\
%ACAT & 14.25 & 14.72 & 14.48 \\
%eASMS & 14.58 & 15.67 & 15.12 \\
%MatFlow & 22.72 & 7.83 & 15.28 \\
%MEEM & 20.19 & 10.50 & 15.35 \\
%HMMTxD & 10.25 & 20.92 & 15.58 \\
%qwsEDFT & 18.07 & 17.75 & 17.91 \\
%ABS & 21.30 & 15.67 & 18.48 \\
\bottomrule
\end{tabular}

}\vspace{1mm}
	\caption{Results for the top 3 trackers on VOT2014. The mean overlap and failure rate is reported in the first two columns. The accuracy rank, robustness rank and the combined final rank are shown in the remaining columns. Our tracker obtains the best performance on this dataset.}\vspace{-3mm}
	\label{tab:VOT_comparison}
\end{table}

\section{Conclusions}
We propose Spatially Regularized Discriminative Correlation Filters (SRDCF) to address the limitations of the standard DCF. The introduced spatial regularization component enables the correlation filter to be learned on larger image regions, leading to a more discriminative appearance model.
By exploiting the sparsity of the regularization operation in the Fourier domain, we derive an efficient optimization strategy for learning the filter. The proposed learning procedure employs the Gauss-Seidel method to solve for the filter in the Fourier domain. We perform comprehensive experiments on four benchmark datasets. Our SRDCF outperforms existing trackers on all four datasets.

\noindent\textbf{Acknowledgments}:
This work has been supported by SSF (CUAS) and VR (VIDI, EMC${}^2$, ELLIIT, and CADICS).

{\small
	\bibliographystyle{ieee}
	\bibliography{RegDCF_tracking}
}

\end{document}